\documentclass[11pt]{article}
\PassOptionsToPackage{table,xcdraw}{xcolor}

\usepackage[preprint]{acl}

\usepackage{times}
\usepackage{latexsym}
\usepackage[T1]{fontenc}
\usepackage[utf8]{inputenc}
\usepackage{microtype}
\usepackage{inconsolata}

\usepackage{amsmath,amsfonts,bm}

\def\eqref#1{equation~\ref{#1}}

\def\1{\bm{1}}

\DeclareMathAlphabet{\mathsfit}{\encodingdefault}{\sfdefault}{m}{sl}
\SetMathAlphabet{\mathsfit}{bold}{\encodingdefault}{\sfdefault}{bx}{n}

\usepackage[table]{xcolor}
\usepackage{booktabs}
\usepackage{booktabs}
\usepackage{multirow}
\usepackage{adjustbox}
\definecolor{oursgreen}{RGB}{232,245,233}
\usepackage{amsmath,amssymb,amsfonts,amsthm}
\usepackage{booktabs,tabularx,array,multirow,makecell}
\usepackage{graphicx}
\usepackage{algorithm,algorithmic}
\usepackage{enumitem,adjustbox,bm,needspace}
\usepackage{caption,subcaption}
\usepackage{placeins}
\usepackage[noabbrev,nameinlink]{cleveref}

\newcommand{\ParaTitle}[1]{\noindent\textbf{#1}}
\title{SwarmRSI: Recursive Self-Improvement \\ through Evolving Scientific Agent Swarm}

\author{%
  \textbf{Qianru Meng\textsuperscript{1} \quad Xi Zhang\textsuperscript{1} \quad Konstantinos Papakostas\textsuperscript{1} \quad Congfeng Cao\textsuperscript{1}} \\
  \textbf{Wenxiang Hu\textsuperscript{1} \quad Tianyi Lu\textsuperscript{2} \quad Xiao Zhang\textsuperscript{3} \quad Lun Zhou\textsuperscript{1}} \\
  \textbf{Zaiqiao Meng\textsuperscript{4} \quad Zhaochun Ren\textsuperscript{2} \quad Xiaohui Yan\textsuperscript{1}\thanks{Corresponding authors.} \quad Yougang Lyu\textsuperscript{1}\footnotemark[1]} \\[0.5em]
  \textsuperscript{1}Huawei Technologies Co., Ltd. \quad \textsuperscript{2}Leiden University \quad \textsuperscript{3}University of Groningen \\
  \textsuperscript{4}Huazhong University of Science and Technology \\
  {\small\texttt{qm17509@gmail.com, yanxiaohui2@huawei.com, youganglyu@gmail.com}}
}

\begin{document}
\raggedbottom
\newlength{\maintextfloatsep}
\newlength{\mainfloatsep}
\newlength{\maindbltextfloatsep}
\newlength{\maindblfloatsep}
\setlength{\maintextfloatsep}{\textfloatsep}
\setlength{\mainfloatsep}{\floatsep}
\setlength{\maindbltextfloatsep}{\dbltextfloatsep}
\setlength{\maindblfloatsep}{\dblfloatsep}
\setlength{\textfloatsep}{18pt plus 2pt minus 2pt}
\setlength{\floatsep}{16pt plus 2pt minus 2pt}
\setlength{\dbltextfloatsep}{18pt plus 2pt minus 2pt}
\setlength{\dblfloatsep}{16pt plus 2pt minus 2pt}
\captionsetup[figure]{skip=7pt}
\setcounter{dbltopnumber}{1}
\maketitle
\begin{abstract}
Recursive self-improvement (RSI) in scientific research aims to improve agents' research capabilities through their own experimental experience.
Scientific agent swarms can pursue this goal when one agent's findings and implementations advance another's research.
Two obstacles stand in the way.
A failed experiment does not reveal whether its hypothesis or its implementation was at fault, and a program without overall gains can hide components that other agents need.
We introduce \textsc{SwarmRSI}, a framework in which a swarm improves its research by evolving the research states its agents inherit, including hypotheses, programs and experimental evidence.
Feedback-Guided Swarm Mutation uses other agents' feedback to decide whether to revise a hypothesis or its program.
Complementary Swarm Recombination tests retained hypotheses on other agents' programs and combines complementary components.
Recursive Swarm Evolution returns every outcome to a shared graph that guides which state each agent updates next.
\textsc{SwarmRSI} achieves state-of-the-art performance
across three benchmarks, scoring 76.2 on OpenProblems,
attaining a 44.4\% Match-SOTA rate on NatureBench,
and earning medals on 81.8\% of MLE-Bench Lite tasks.
Across 90 NatureBench tasks, 77\% of the best programs
descend from feedback-guided mutation or recombination.
Biologists further validate two generated programs
in studies of intestinal and splenic immunity.
The code is available\footnote{\url{https://github.com/EvoScientist/SwarmRSI}}.
\end{abstract}

\section{Introduction}
\label{sec:introduction}

\suppressfloats[t]
\begin{figure}[t]
\centering
\includegraphics[width=\columnwidth]{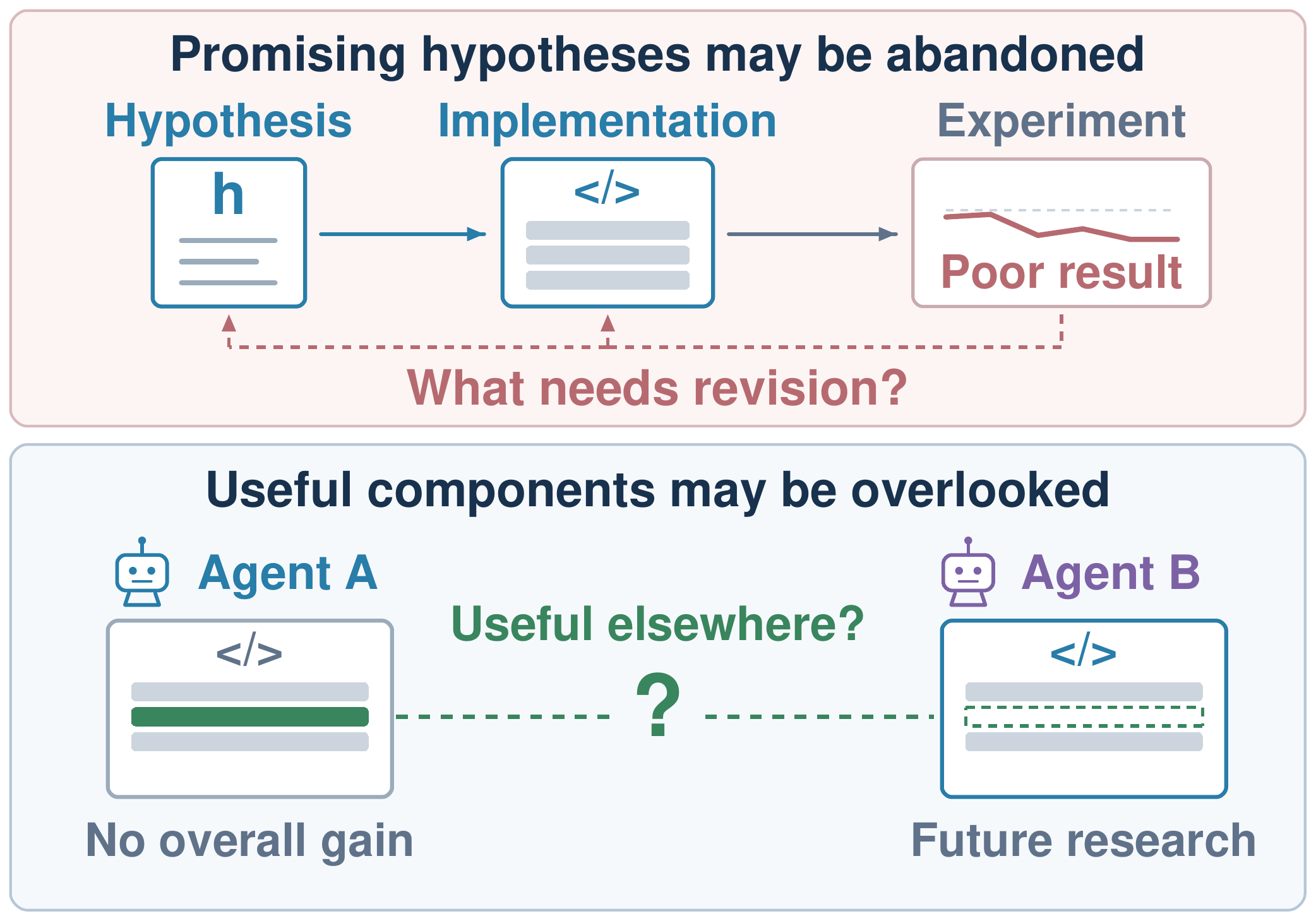}
\caption{Two limitations motivating \textsc{SwarmRSI}. In the top example, a poor result does not show whether the hypothesis or its implementation needs revision, and a promising hypothesis is abandoned. In the bottom example, a program without overall gains holds components useful to another agent's later research. Dashed links mark open revisions and potential reuse.}
\label{fig:problem_statement}
\end{figure}

Recursive self-improvement (RSI) in scientific research aims to improve agents' research capabilities through their own experimental experience~\citep{guo2026aqua,srikanth2026recursive}.
Scientific agent swarms offer a way to pursue this goal by coordinating idea generation, implementation, and evaluation across specialized agents~\citep{yang2025rdagent,tang2025airesearcher,yamada2025aiscientistv2,lee2026mass}.
Persistent memories and shared histories retain experimental experience for later research~\citep{evoscientist2026,liu2026mlmaster2,du2026mlevolve,weng2026deepscientist}, and accumulated histories can also support improvements to exploration strategies~\citep{zheng2026dreamrsi}.
Together, these capabilities create an opportunity for agents exploring different directions to use one another's findings and implementations in subsequent research.

However, sharing experience does not ensure that agents use it to make better research decisions.
Program evolution refines executable candidates through execution feedback~\citep{jiang2025aide,nam2025mlestar,novikov2025alphaevolve,liu2025mlmaster}, while hypothesis evolution improves research directions through iterative generation and evaluation~\citep{gottweis2026coscientist,wang2026evodiverse,liu2026hypoevolve,jin2026arbor}.
As \Cref{fig:problem_statement} illustrates, promising hypotheses are abandoned after inadequate implementations, and useful components are overlooked in programs that perform poorly overall.

\Needspace{2\baselineskip}
We hypothesize that these limitations arise from two interrelated dependencies in scientific research. The value of a hypothesis depends on how it is implemented and how rigorously it is tested.
An unsuccessful experiment therefore leaves unclear whether the research direction or its implementation needs revision~\citep{ma2026sage,koo2026hypoexplore}, and feedback from other agents identifies which changes are worth pursuing.
Intermediate results also become stepping stones for other search paths~\citep{wang2019poet,wang2020enhancedpoet}, and their value depends on combinations with later findings or implementations.
Their current performance thus underestimates their contribution to other agents' future research.
This leads to our central question of how scientific agent swarms can evolve toward recursive self-improvement.

To address this, we introduce \textsc{SwarmRSI}, a framework that evolves scientific agent swarms for recursive self-improvement.
Each agent uses its hypotheses, programs, experimental evidence, and unresolved needs to plan further research, and a shared graph records these inheritable research states.
In Feedback-Guided Swarm Mutation, agents revise hypotheses and programs using other agents' experimental feedback.
Performance differences inform hypothesis revision, and unmet research needs guide implementation changes.
In Complementary Swarm Recombination, agents revisit retained hypotheses on other agents' implementations and combine program components with complementary strengths.
Recursive Swarm Evolution evaluates the resulting candidates, retains useful results, feedback and unresolved needs in the shared graph, and uses these states to decide which hypotheses agents revisit and which programs they refine.
Each experiment thus updates the states that later agents start from.

We evaluate \textsc{SwarmRSI} on OpenProblems~\citep{luecken2025openproblems}, NatureBench~\citep{wang2026naturebench} and MLE-Bench Lite~\citep{chan2025mlebench}, 118 scientific coding and machine-learning engineering tasks, with two backbones.
It achieves SOTA results on OpenProblems and NatureBench and outperforms hypothesis and program evolution agents with the same backbone.
With DeepSeek-V4-Pro in one seed over all 90 NatureBench tasks, about 77\% of its final best programs descend from a rebase, a crossover or an update triggered by feedback.
Biologists reproduce two of its programs and use them in follow-up analyses of intestinal and splenic immunity.

\begin{figure*}[!t]
\centering
\includegraphics[width=\linewidth]{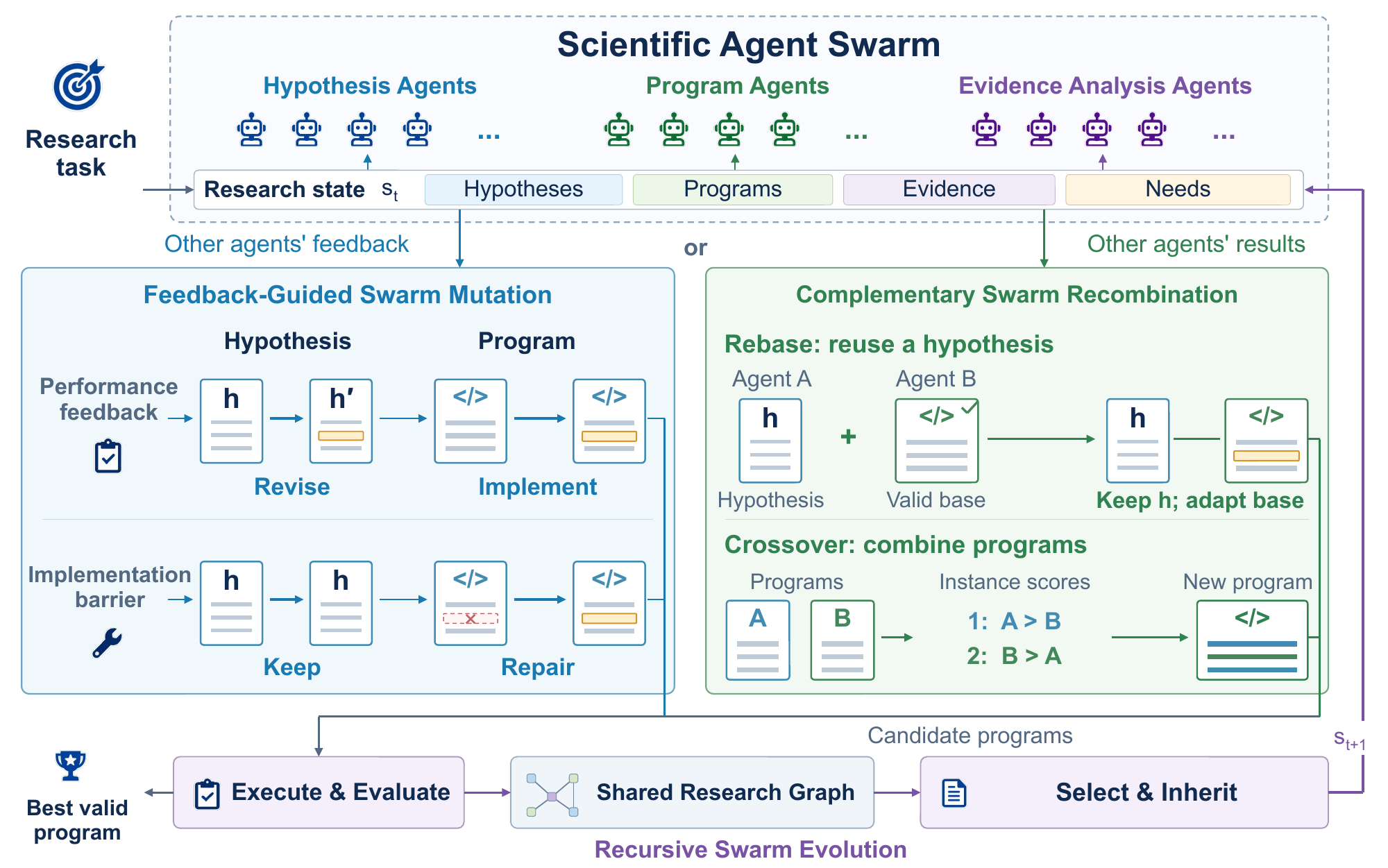}
\caption{Overview of \textsc{SwarmRSI}. Scientific agents use research states assembled from a shared graph. Feedback-Guided Swarm Mutation revises a hypothesis and implements the revision, or retains the hypothesis and repairs its program. Highlighted lines show local modifications. Complementary Swarm Recombination preserves a hypothesis while adapting another agent's valid program base (rebase), or compares results across instances and adapts components from complementary programs into one new program (crossover). Blue and green code lines indicate crossover sources. The comparisons and code layouts are schematic. All candidates undergo execution and evaluation, and all attempts, whether successful, failed or without gain, update the graph and the next agent's state $s_{t+1}$.}
\label{fig:architecture}
\end{figure*}

\Needspace{4\baselineskip}
Our main contributions are:

\begin{itemize}[leftmargin=*,nosep]

  \item We formulate RSI for scientific agent swarms as evolving inheritable research states, so that each experiment updates the hypotheses, programs and evidence that later agents start from.

  \Needspace{2\baselineskip}
  \item We introduce \textsc{SwarmRSI}, which updates these states through Feedback-Guided Swarm Mutation, Complementary Swarm Recombination and Recursive Swarm Evolution.

  \Needspace{2\baselineskip}
  \item \textsc{SwarmRSI} achieves SOTA results on OpenProblems and NatureBench, most of its final NatureBench programs build on inherited states, and biologists reproduce its programs for immune analyses. 
 
\end{itemize}

\section{Related Work}
\label{sec:related_work}

\subsection{LLM Agents for Scientific Research}
\label{sec:rw_scientific_agents}
LLM agents automate scientific research through hypothesis generation, experiment implementation, and result evaluation~\citep{tang2025airesearcher,yamada2025aiscientistv2}.
Agents centred on programs search over or refine code with execution feedback~\citep{jiang2025aide,nam2025mlestar}, R\&D-Agent coordinates proposal generation and implementation through specialized roles~\citep{yang2025rdagent}, and systems with memory carry execution experience into later experiments~\citep{liu2026mlmaster2,du2026mlevolve}.
Closest to our setting, Arbor links explicit hypotheses to programs and experimental evidence in a persistent hypothesis tree~\citep{jin2026arbor}, and SAGE evaluates competing failure explanations to direct interventions toward hypotheses, experimental design, or implementation~\citep{ma2026sage}.
\textsc{SwarmRSI} likewise grounds hypotheses in experiments and lets agents use one another's feedback and partial results.

\subsection{Recursive Self-Improvement of Agents}
\label{sec:rw_rsi}
Recursive self-improvement uses an agent's own experience to improve the mechanisms that guide its subsequent activity.
AIDE$^2$ retains accepted code modifications as the basis for further self-improvement~\citep{srikanth2026recursive}, Dream-RSI refines exploration policies with replay simulators built from historical discovery trees~\citep{zheng2026dreamrsi}, MASS alternates workflow evolution with model fine-tuning~\citep{lee2026mass}, and AQuA uses recorded experiments to guide later research proposals~\citep{guo2026aqua}.
These methods improve the agent's code, policy, workflow, or proposals.
\textsc{SwarmRSI} instead evolves the research states that agents inherit, revising them with other agents' feedback and reusing retained hypotheses and program components across agents.

\section{SwarmRSI}
\label{sec:method}

\textsc{SwarmRSI} evolves scientific agent swarms by updating the research states agents inherit.
Feedback-Guided Swarm Mutation uses other agents' feedback to guide revisions, and Complementary Swarm Recombination reuses their hypotheses and program components.
Recursive Swarm Evolution evaluates every candidate and retains its outcome, so each experiment informs subsequent research.
\Cref{fig:architecture} shows the shared research state and the execution loop that updates it.

\subsection{Preliminaries}
\label{subsec:preliminaries}

\ParaTitle{Scientific research task.}
Given a research goal $q$, a starter program $p_0$, and budget $B$, we seek a valid program with high utility.
A hypothesis $h$ specifies a modeling or algorithmic mechanism, and a program $p$ implements it under the task's constraints.
Different programs may test the same hypothesis.
Let $\mathcal P_{\mathrm{valid}}(B)$ contain the valid programs found and evaluated within budget.
The system returns
\begin{equation}
 p_B^\star \in \operatorname*{arg\,max}_{p\in\mathcal P_{\mathrm{valid}}(B)} u(p),
 \label{eq:research_objective}
\end{equation}
where $u(p)$ is the utility measured under the task's evaluation protocol.

\ParaTitle{Scientific agent swarm.}
A scientific agent is a research policy for hypothesis proposal, program development, or evidence analysis.
At research step $t$, the acting agent uses a research state
\begin{equation}
 s_t=(\mathcal H_t,\mathcal P_t,\mathcal E_t,\mathcal U_t),
 \label{eq:research_state}
\end{equation}
containing the hypotheses, programs, experimental evidence, and unresolved needs available for its current action.
It assembles this state from relevant records in the shared graph $\mathcal G_t$, including other agents' results, and unresolved research needs.
Each record $\xi_i=(H_i,P_i,X_i,E_i)$ contains the tested hypothesis, parent and resulting programs, outcome, and execution and implementation evidence.
The graph retains failed attempts and attempts without gain, and its edges track hypothesis--program associations and parent or reuse relations.
Mutation and recombination use $s_t$ to generate subsequent candidates.

\subsection{Feedback-Guided Swarm Mutation}
\label{subsec:swarm_mutation}

To guide revisions of hypotheses and their implementations, Feedback-Guided Swarm Mutation uses experimental feedback from the swarm.
An unsuccessful run points to an implementation barrier, and differences across instances point to a revised research direction.
At step $t$, the agent uses the selected parent pair $(h_t,p_t)$ and state $s_t$, which provides outcomes, implementation evidence, and unresolved needs from its own and other agents' records.

The agent proposes a hypothesis and a revision plan:
\begin{equation}
 (\tilde h_t,\delta_t)=\operatorname{Propose}(q,h_t,p_t;s_t),
 \label{eq:mutation_proposal}
\end{equation}
where $\delta_t$ describes the intended changes and their rationale.
Differences across instances guide hypothesis revisions, and execution failures or unrealized interventions guide program repair.
We call a revision that changes the hypothesis an \emph{H-mutation}, and one that keeps it ($\tilde h_t=h_t$) and repairs only the implementation a \emph{P-mutation}.
The agent then implements the plan on the parent program using relevant state information:
\begin{equation}
 \tilde p_t=\operatorname{Implement}(p_t,\tilde h_t,\delta_t;s_t).
 \label{eq:mutation_implementation}
\end{equation}

The resulting candidate $(\tilde h_t,\tilde p_t)$ retains its parent relation and revision rationale in the shared research state.
It then enters the shared evaluation loop, which tests validity and measures utility.

\subsection{Complementary Swarm Recombination}
\label{subsec:swarm_recombination}

Complementary Swarm Recombination reuses partial results across agents, including programs whose overall scores did not improve.

\ParaTitle{Hypothesis reuse.}
For a hypothesis $h_t$ with unresolved implementation needs, an agent selects a valid base $b_t$ from another agent by matching these needs to program mechanisms~\citep{wiegand2001collaboration}.
The \emph{rebase} operation keeps $h_t$ and adapts its implementation to $b_t$, giving $(\tilde h_t,\tilde p_t)=(h_t,\operatorname{Rebase}(h_t,b_t;s_t))$.

\ParaTitle{Program recombination.}
\emph{Crossover} combines programs with complementary strengths across instances~\citep{holland1975adaptation,goldberg1989genetic}.
For each dataset or subtask $k$, let $p_k^\star$ be the valid program with the highest utility $u_k$ observed by step $t$.
Donors are the programs that outperform the parent on some instance, ranked by their largest advantage,
\begin{equation}
 \begin{aligned}
 \mathcal D(p_t)&=\{p_k^\star:u_k(p_k^\star)>u_k(p_t)\},\\
 \Delta_t(d)&=\max_{k:p_k^\star=d}\bigl[u_k(d)-u_k(p_t)\bigr].
 \end{aligned}
 \label{eq:instance_sources}
\end{equation}
Each instance takes its components from the highest ranked donor that outperforms the parent there and keeps the parent's components elsewhere.
Both operations record their sources and send the new candidate to the shared evaluation loop.

\begin{table*}[!t]
\centering
\footnotesize
\captionsetup{font=footnotesize,skip=3pt}
\setlength{\heavyrulewidth}{0.7pt}
\setlength{\lightrulewidth}{0.4pt}
\setlength{\cmidrulewidth}{0.4pt}
\setlength{\tabcolsep}{2pt}
\renewcommand{\arraystretch}{1.08}
\begin{tabularx}{\textwidth}{ll*{7}{>{\centering\arraybackslash}X}}
\toprule
Method & Backbone & Denoise & Label & Batch & SVG & Decomp & Visual & Avg $\uparrow$\\
\midrule
Human Expert & Task-specific
& 28.0 & \underline{85.0} & 71.0 & 66.0 & 49.0 & 44.0 & 57.2\\
Biomni & Claude 4 Sonnet
& 16.0 & \textbf{89.0} & 82.0 & 16.0 & 53.0 & 35.0 & 48.5\\
ChatGPT & GPT-5
& 3.0 & 81.0 & \textbf{83.0} & 60.0 & 74.0 & 38.0 & 56.5\\
AIDE$^{\dagger}$ & DeepSeek-V4-Pro
& 27.4 & 70.6 & 78.0 & 72.8 & 46.1 & 39.8 & 55.8\\
TusoAI$^{\dagger}$ & DeepSeek-V4-Pro
& 31.5 & 58.7 & 79.5 & 79.0 & \underline{84.1} & 69.8 & 67.1\\
\midrule
\rowcolor{oursgreen}
& DeepSeek-V4-Pro
& \underline{33.6} & 74.2 & \underline{82.5} & \underline{83.9} & \textbf{86.0} & \underline{84.8} & \underline{74.2}\\
\rowcolor{oursgreen}
\multicolumn{1}{c}{\multirow{-2}{*}{\textbf{\textsc{SwarmRSI}}$^{\dagger}$}}
& Gemini 3.1 Pro Preview
& \textbf{40.2} & 80.7 & 78.8 & \textbf{87.8} & 83.4 & \textbf{86.4} & \textbf{76.2}\\
\bottomrule
\end{tabularx}
\caption{Main results on the six OpenProblems tasks. Scores are normalized and scaled by 100. Avg is the six-task mean, recomputed from reported task scores for published references. $\dagger$ denotes our means over three seeds. \textbf{Bold} and \underline{underlined} values indicate best and second best results, including ties at the reported precision.}
\label{tab:openproblems_results}
\par\vspace{7pt}
\setlength{\tabcolsep}{1pt}
\begin{tabularx}{\textwidth}{ll*{14}{>{\centering\arraybackslash}X}}
\toprule
 & & \multicolumn{2}{c}{All}
 & \multicolumn{2}{c}{Protein}
 & \multicolumn{2}{c}{Cellular}
 & \multicolumn{2}{c}{Physical}
 & \multicolumn{2}{c}{Molec.}
 & \multicolumn{2}{c}{Relat.}
 & \multicolumn{2}{c}{Biomed.}\\
\cmidrule(lr){3-4}\cmidrule(lr){5-6}
\cmidrule(lr){7-8}\cmidrule(lr){9-10}
\cmidrule(lr){11-12}\cmidrule(lr){13-14}
\cmidrule(lr){15-16}
Method & Backbone
 & S$\uparrow$ & M$\uparrow$
 & S$\uparrow$ & M$\uparrow$
 & S$\uparrow$ & M$\uparrow$
 & S$\uparrow$ & M$\uparrow$
 & S$\uparrow$ & M$\uparrow$
 & S$\uparrow$ & M$\uparrow$
 & S$\uparrow$ & M$\uparrow$\\
\midrule
\multirow{5}{*}{Claude Code}
 & DeepSeek-V4-Pro
 & 4.4 & 23.3 & \underline{6.3} & \underline{37.5} & 9.7 & 29.0 & 0.0 & 7.7 & \underline{0.0} & 9.1 & \underline{0.0} & \underline{60.0} & 0.0 & 7.1\\
 & MiniMax-M3
 & 6.7 & 27.8 & \underline{6.3} & \underline{37.5} & 12.9 & 29.0 & 7.7 & 15.4 & \underline{0.0} & 36.4 & \underline{0.0} & \underline{60.0} & 0.0 & 7.1\\
 & GLM-5.2
 & 13.3 & 35.6 & \textbf{12.5} & \textbf{43.8} & \underline{22.6} & \underline{41.9} & 15.4 & 15.4 & \underline{0.0} & 36.4 & \underline{0.0} & \underline{60.0} & 7.1 & 21.4\\
 & Kimi K3
 & 14.4 & 33.3 & \underline{6.3} & 31.3 & \underline{22.6} & 38.7 & 15.4 & 15.4 & \textbf{9.1} & 36.4 & \underline{0.0} & \underline{60.0} & \underline{14.3} & \underline{28.6}\\
 & Qwen 3.8 Max 0902
 & \textbf{17.8} & 35.6 & \textbf{12.5} & \underline{37.5} & \textbf{25.8} & 38.7 & \underline{23.1} & 23.1 & \underline{0.0} & 27.3 & \underline{0.0} & \underline{60.0} & \textbf{21.4} & \textbf{35.7}\\
\multirow{2}{*}{Codex}
 & GPT-5.4
 & 6.7 & 24.4 & \underline{6.3} & \underline{37.5} & 9.7 & 22.6 & 15.4 & 23.1 & \underline{0.0} & 18.2 & \underline{0.0} & \underline{60.0} & 0.0 & 7.1\\
 & GPT-5.5
 & 4.4 & 25.6 & \underline{6.3} & 31.3 & 9.7 & 25.8 & 0.0 & 7.7 & \underline{0.0} & 18.2 & \underline{0.0} & \underline{60.0} & 0.0 & \underline{28.6}\\
Gemini CLI
 & Gemini 3.5 Flash
 & 10.0 & 25.6 & \underline{6.3} & 31.3 & 16.1 & 32.3 & \underline{23.1} & 23.1 & \underline{0.0} & 18.2 & \underline{0.0} & \underline{60.0} & 0.0 & 0.0\\
Arbor$^{\dagger}$
 & DeepSeek-V4-Pro
 & 12.2 & \underline{36.7$^{*}$} & \underline{6.3} & 31.3 & 16.1 & \textbf{48.4} & \underline{23.1} & 23.1 & \underline{0.0} & 27.3 & \underline{0.0} & \underline{60.0} & \underline{14.3} & \underline{28.6}\\
AIDE$^{\dagger}$
 & DeepSeek-V4-Pro
 & 10.0 & 27.8$^{*}$ & \underline{6.3} & 25.0 & 16.1 & 35.5 & 15.4 & 15.4 & \underline{0.0} & 27.3 & \underline{0.0} & \underline{60.0} & 7.1 & 14.3\\
\midrule
\rowcolor{oursgreen}
 & DeepSeek-V4-Pro
 & \underline{15.6} & \textbf{44.4} & 0.0 & \underline{37.5} & \underline{22.6} & \textbf{48.4} & \textbf{53.8} & \underline{53.8} & \underline{0.0} & \underline{45.5} & \underline{0.0} & \underline{60.0} & 0.0 & \underline{28.6}\\
\rowcolor{oursgreen}
\multicolumn{1}{c}{\multirow{-2}{*}{\textbf{\textsc{SwarmRSI}}$^{\dagger}$}}
 & Gemini 3.1 Pro Preview
 & 14.4 & \textbf{44.4} & \underline{6.3} & \textbf{43.8} & \underline{22.6} & 38.7 & \underline{23.1} & \textbf{61.5} & \textbf{9.1} & \textbf{54.5} & \textbf{20.0} & \textbf{80.0} & 0.0 & 21.4\\
\bottomrule
\end{tabularx}
\caption{Main results on all 90 NatureBench tasks. S and M denote Surpass-SOTA and Match-SOTA rates (\%). $\dagger$ denotes our means over three seeds. $*$ marks All-M differences from \textsc{SwarmRSI} with the same backbone significant at $p<0.05$ (McNemar test). \textbf{Bold} and \underline{underlined} values indicate best and second best rates, including ties.}
\label{tab:naturebench_results}
\end{table*}

\begin{table*}[!t]
\centering
\footnotesize
\captionsetup{font=footnotesize,skip=3pt}
\setlength{\heavyrulewidth}{0.7pt}
\setlength{\lightrulewidth}{0.4pt}
\setlength{\cmidrulewidth}{0.4pt}
\renewcommand{\arraystretch}{1.08}
\setlength{\tabcolsep}{2.5pt}

\begin{tabularx}{\textwidth}{ll>{\centering\arraybackslash}Xccc>{\centering\arraybackslash}X}
\toprule
Method & Backbone
& \mbox{Above Median $\uparrow$}
& Bronze
& Silver
& Gold $\uparrow$
& \mbox{Any-Medal $\uparrow$}\\
\midrule

\multirow[c]{2}{*}{MLEvolve}
& DeepSeek-V4-Pro$^{\dagger}$
& 81.8 & 9.1 & 27.3 & 36.4 & 72.7\\
& Gemini 3.1 Pro Preview
& -- & -- & -- & -- & 80.3\\

\multirow[c]{2}{*}{Arbor}
& DeepSeek-V4-Pro$^{\dagger}$
& 86.4 & 4.5 & 31.8 & 40.9 & 77.3\\
& GPT-5.5
& \textbf{95.5} & 0.0 & 9.1 & \textbf{77.3} & \textbf{86.4}\\

ML-Master 2.0
& \mbox{DeepSeek-V3.2-Speciale}
& 84.9 & 13.6 & 31.8 & 30.3 & 75.8\\

Famou-Agent 2.0
& Gemini 3 Pro Preview
& \underline{93.9} & 12.1 & 18.2 & \underline{50.0} & 80.3\\

\midrule

\rowcolor{oursgreen}
& DeepSeek-V4-Pro
& 90.9 & 9.1 & 27.3 & 45.5 & \underline{81.8}\\

\rowcolor{oursgreen}
\multicolumn{1}{c}{\multirow{-2}{*}{\textbf{\textsc{SwarmRSI}}$^{\dagger}$}}
& Gemini 3.1 Pro Preview
& 86.4 & 9.1 & 31.8 & 40.9 & \underline{81.8}\\

\bottomrule
\end{tabularx}

\caption{Main results on 22 MLE-Bench Lite tasks (\%). $\dagger$ denotes our means over three seeds. Any-Medal combines exclusive medal rates before rounding, and -- denotes unreported values. \textbf{Bold} and \underline{underlined} values indicate best and second best Above Median, Gold, and Any-Medal rates, including ties.}
\label{tab:mle_results}
\end{table*}

\subsection{Recursive Swarm Evolution}
\label{subsec:recursive_swarm_evolution}

Recursive Swarm Evolution decides which update to make next and returns every outcome to the shared graph.
It maintains a hypothesis frontier $\mathcal F_H$ of active hypotheses and a program frontier $\mathcal F_P$ of valid program representatives. For \emph{H/P selection}, which decides whether the next step revises a hypothesis (H) or a program (P), it scores each feasible frontier $f\in\{H,P\}$ with a discounted-UCB index~\citep{auer2002finite,garivier2011upper}
\begin{equation}
 S_f=\hat\mu_f+\beta\sqrt{\frac{\log(1+N_H+N_P)}{N_f}}+\gamma\psi_f,
 \label{eq:frontier_score}
\end{equation}
where $\hat\mu_f$ is recent improvement per token cost, $N_f$ is the discounted experiment count, and $\psi_f$ is the saturated pressure from unresolved needs that $f$ can address.
The weights $\beta$ and $\gamma$ set exploration and request pressure.
Untried frontiers and frontiers that have waited long take priority over the score.

Each candidate is executed and evaluated under the task protocol.
Execution checks test completion and task compliance, and source and output comparisons check whether the intended change was implemented.
The record $\xi_t$ of every attempt, including failed ones and those without gain, updates the graph to $\mathcal G_{t+1}$, from which the next acting agent assembles its state $s_{t+1}$.
An implementation failure keeps the hypothesis and requests program repair, mixed gains and losses across instances request a revised hypothesis, and a new incumbent requests that retained hypotheses be realized on it.
The loop repeats until budget $B$ is exhausted and returns the best valid program.

\section{Experiments}
\label{sec:experiments}
\label{subsec:setup}

\subsection{Research Questions}
\label{subsec:research_questions}
We address four research questions and illustrate biological applications through case studies:
\textbf{RQ1}: How does \textsc{SwarmRSI} perform across benchmarks and backbones, and at what token cost?
\textbf{RQ2}: How do Feedback-Guided Swarm Mutation, Complementary Swarm Recombination, and Recursive Swarm Evolution contribute to performance?
\textbf{RQ3}: How do experimental feedback and reused intermediate results guide the swarm's subsequent research within and across tasks?
\textbf{RQ4}: What are the main failure modes, and how frequent are they?

\begin{figure}[!t]
\centering
\includegraphics[width=\columnwidth]{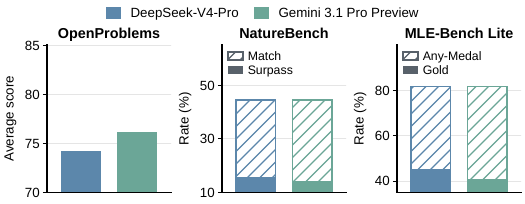}
\caption{\textsc{SwarmRSI} with two backbones. Full bars give Match-SOTA and Any-Medal rates (\%), and solid portions give Surpass-SOTA and Gold rates.}
\label{fig:backbone_comparison}
\end{figure}

\begin{table}[t]
\centering
\footnotesize
\setlength{\tabcolsep}{3pt}
\begin{tabular*}{\columnwidth}{@{\extracolsep{\fill}}lccc@{}}
\toprule
Variant & Match & Surpass & Mean $g$\\
\midrule
\textbf{\textsc{SwarmRSI}} & \textbf{44.4} & \textbf{15.6} & \textbf{12.8}\\
\addlinespace[2pt]
\multicolumn{4}{@{}l}{\textit{Feedback-Guided Swarm Mutation}}\\
\hspace{0.3em}w/o H-mutation& 34.4 & 8.9 & 7.2\\
\hspace{0.3em}w/o P-mutation& 31.1 & 6.7 & 5.4\\
\addlinespace[2pt]
\multicolumn{4}{@{}l}{\textit{Complementary Swarm Recombination}}\\
\hspace{0.3em}w/o recombination& 37.8 & 7.8 & 7.6\\
\addlinespace[2pt]
\multicolumn{4}{@{}l}{\textit{Recursive Swarm Evolution}}\\
\hspace{0.3em}w/o shared graph& 36.7 & 10.0 & 8.1\\
\hspace{0.3em}w/o H/P selection& 35.6 & 8.9 & 4.6\\
\bottomrule
\end{tabular*}
\caption{NatureBench ablations on all 90 tasks with DeepSeek-V4-Pro, averaged over three seeds and grouped by the mechanism each variant removes. Match and Surpass are Match-SOTA and Surpass-SOTA rates (\%), and mean $g$ is scaled by 100.}
\label{tab:ablation}
\end{table}

\subsection{Benchmarks}
\label{subsec:benchmarks}
We evaluate \textsc{SwarmRSI} on scientific coding and machine-learning engineering tasks across three benchmarks. \textbf{OpenProblems}~\citep{luecken2025openproblems} contains six single-cell analysis tasks following TusoAI~\citep{turcan2026tusoai}. \textbf{NatureBench}~\citep{wang2026naturebench} contains all 90 scientific coding tasks across six domains. \textbf{MLE-Bench Lite}~\citep{chan2025mlebench} contains 22 competitions from its low-complexity split.

\subsection{Baselines}
\label{subsec:baselines}
We compare against \textbf{program evolution agents} (AIDE and MLEvolve), \textbf{hypothesis evolution agents} (Arbor), \textbf{domain-specific scientific agents} (TusoAI and Biomni~\citep{huang2026biomni}), and \textbf{general coding agents} (Claude Code~\citep{anthropic2026claudecode}, Codex~\citep{openai2026codexcli} and Gemini CLI~\citep{google2026geminicli}).

\subsection{Evaluation Metrics}
\label{subsec:metrics}
OpenProblems reports normalized task scores scaled by 100 and their unweighted mean.
NatureBench reports Match-SOTA ($g\geq0$) and Surpass-SOTA ($g>0.1$) rates, where $g$ is the official direction-adjusted improvement over SOTA.
MLE-Bench Lite reports Above Median and medal rates from official grading on the private test split.
Rates retain all tasks in the denominator, including invalid or missing results.
Match-SOTA comparisons with Arbor and AIDE on NatureBench use 95\% paired bootstrap confidence intervals and McNemar tests, and other comparisons are descriptive.
Token cost sums uncached input and output tokens over all LLM calls in a search.

\subsection{Implementation Details}
\label{subsec:implementation}
\textsc{SwarmRSI} and rerun baselines use DeepSeek-V4-Pro~\citep{deepseek2026v4} under matched settings for each benchmark.
Results from our runs in Tables~\ref{tab:openproblems_results}--\ref{tab:mle_results} are averaged over three random seeds.
Published results keep the settings of their sources.
We additionally evaluate \textsc{SwarmRSI} with Gemini~3.1 Pro Preview~\citep{google2026gemini31pro}.
In each run, all agents share one backbone and advance lineages that exchange results through the shared graph.

\section{Results and Analysis}
\label{sec:results}
To answer our research questions, we compare benchmark performance, conduct component ablations, examine search trajectories, analyze annotated failures, and report biological case studies.

\subsection{Main Results (RQ1)}
\label{subsec:main}
\Cref{tab:openproblems_results,tab:naturebench_results,tab:mle_results} summarize benchmark performance, and we make three observations.

\begin{itemize}[leftmargin=*,nosep]
  \item \textbf{\textsc{SwarmRSI} outperforms domain-specific and general coding agents.} With DeepSeek-V4-Pro, it reaches an OpenProblems average of 74.2 against 67.1 for TusoAI and exceeds Claude Code with the same backbone by 21.1 percentage points (pp) in NatureBench Match-SOTA.

  \item \textbf{\textsc{SwarmRSI} outperforms hypothesis and program evolution agents.} With the same backbone, its NatureBench Match-SOTA rate exceeds Arbor by 7.8 pp and AIDE by 16.7 pp, both significantly. Its 81.8\% Any-Medal rate on MLE-Bench Lite also exceeds Arbor and MLEvolve, and, as shown in \Cref{app:performance_cost}, it uses fewer tokens than TusoAI and Arbor.

  \item \textbf{\textsc{SwarmRSI} remains strong across backbones.} As shown in \Cref{fig:backbone_comparison}, with Gemini~3.1 Pro Preview it reaches the highest OpenProblems average of 76.2 and the same 44.4\% NatureBench Match-SOTA rate as with DeepSeek-V4-Pro, which exceeds every compared method, and it keeps the 81.8\% Any-Medal rate on MLE-Bench Lite.
\end{itemize}

\subsection{Ablation Studies (RQ2)}
\label{subsec:ablation}
\Cref{tab:ablation} compares \textsc{SwarmRSI} with five variants on all 90 NatureBench tasks using DeepSeek-V4-Pro, grouped by the mechanism they remove and otherwise unchanged.
\textbf{w/o H-mutation} only repairs programs under existing hypotheses, \textbf{w/o P-mutation} only revises hypotheses, \textbf{w/o recombination} disables rebase and crossover, \textbf{w/o shared graph} removes agents' access to one another's records, and \textbf{w/o H/P selection} replaces adaptive H/P selection with fixed H/P alternation.
As shown in \Cref{tab:ablation_ci}, all variants lower mean $g$ with paired 95\% CIs excluding zero, and we make three observations.

\begin{itemize}[leftmargin=*,nosep]
  \item \textbf{Feedback-Guided Swarm Mutation needs both revision modes.} Removing H- or P-mutation lowers Match-SOTA by 10.0 and 13.3 pp, respectively. The two modes are complementary, and program repair contributes more.

  \item \textbf{Complementary Swarm Recombination improves solution quality.} Disabling rebase and crossover lowers Match-SOTA by 6.7 pp and Surpass-SOTA from 15.6\% to 7.8\%, and its drop in mean $g$ excludes zero. RQ3 traces this gain to reused hypotheses and program components in most final programs.

  \item \textbf{Recursive Swarm Evolution depends on the shared graph and H/P selection.} Match-SOTA drops by 7.8 pp without the shared graph, through which agents use other agents' records, and by 8.9 pp without H/P selection, which also causes the largest drop in mean $g$.
\end{itemize}

\subsection{Trajectory Analysis (RQ3)}
\label{subsec:analysis}

\begin{figure}[t]
\centering
\includegraphics[width=\columnwidth]{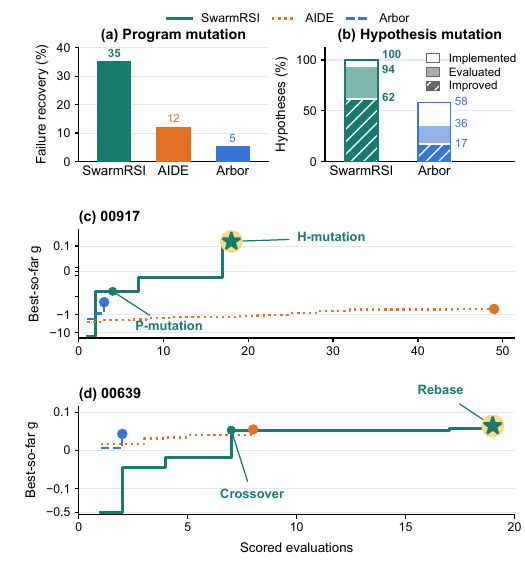}
\caption{Search behaviour on NatureBench with DeepSeek-V4-Pro. (a) Failure recovery, the share of failed implementations whose later refinement yields a new best, over 90 tasks. (b) Nested shares of proposed hypotheses that are implemented (white), evaluated (light) and improved (hatched). (c, d) Best-so-far $g$ of one recorded run per method on symmetric log axes.}
\label{fig:search_diagnostics}
\label{fig:trajectories}
\end{figure}

We examine how \textsc{SwarmRSI} uses feedback and retained research states during search. \Cref{fig:search_diagnostics}a,b compares diagnostics with AIDE and Arbor on all 90 NatureBench tasks, and \Cref{fig:trajectories}c,d shows one run per method on two tasks.

\begin{itemize}[leftmargin=*,nosep]
  \Needspace{4\baselineskip}
  \item \textbf{\textsc{SwarmRSI} recovers more failed implementations and turns more of its hypotheses into improvements.}
  Later refinement turns 35\% of failed implementations into new best valid programs, compared with 12\% for AIDE and 5\% for Arbor. Of its proposed hypotheses, 62\% lead to an improvement, against 17\% for Arbor.
  \Needspace{4\baselineskip}
  \item \textbf{\textsc{SwarmRSI} revises with prior feedback.}
  In one NatureBench seed with DeepSeek-V4-Pro, updates triggered by performance feedback or unresolved implementation requests reach a 28.4\% rate of new bests, against 21.9\% over all updates. On 00917, after nine candidates without improvement, two successive hypothesis mutations revise the program's readout and reach Surpass-SOTA.
  \Needspace{4\baselineskip}
  \item \textbf{\textsc{SwarmRSI} reuses partial results across research trajectories.}
  In the same seed, about 77\% of final best programs descend from at least one rebase, crossover or update triggered by feedback, and these updates yield new bests more often than other updates at every stage of search. On 00639, crossover combines components from three programs to reach Match-SOTA, and a later rebase step activates retained message-passing and ranking-loss code that the program had not executed, further improving $g$. On 00795, rebase tests an unrealized hypothesis on a later program, and crossover then reuses a component from a program that had not improved overall.
  \Needspace{4\baselineskip}
  \item \textbf{Inherited research states transfer to new tasks.}
  As shown in \Cref{app:cross_task}, the state inherited after three generations of source tasks raises mean $g$ on 12 held-out NatureBench tasks from 12.01 to 21.89 and Match-SOTA from 41.7\% to 86.1\% over cold start.
\end{itemize}
\subsection{Failure Analysis (RQ4)}
\label{subsec:failure_analysis}
We annotate 644 failed attempts from 35 \textsc{SwarmRSI} task trajectories that fall below Match-SOTA on NatureBench or earn no medal on MLE-Bench Lite. Two annotators judge implementation fidelity, execution success and improvement over the preceding best valid result, with Cohen's $\kappa=0.92$. \textbf{Most failures are faithful interventions without gains.}
They account for 93.3\% of these failures, and implementation or execution problems for 6.7\%.
A valid score can hide such a problem, as on task 00838, where downstream code overwrites a proposed amino-acid substitution that the evaluation thus never tests, which is why Feedback-Guided Swarm Mutation reads implementation evidence alongside scores. On task 02316, by contrast, a negative-binomial mixture with a Potts spatial prior runs correctly but collapses all locations into one cluster, a faithful intervention without gain.

\setlength{\textfloatsep}{24pt plus 2pt minus 2pt}
\subsection{Biological Case Studies}
\label{subsec:case_study}

\begin{figure}[!t]
\centering
\captionsetup{skip=10pt}
\includegraphics[width=\linewidth]{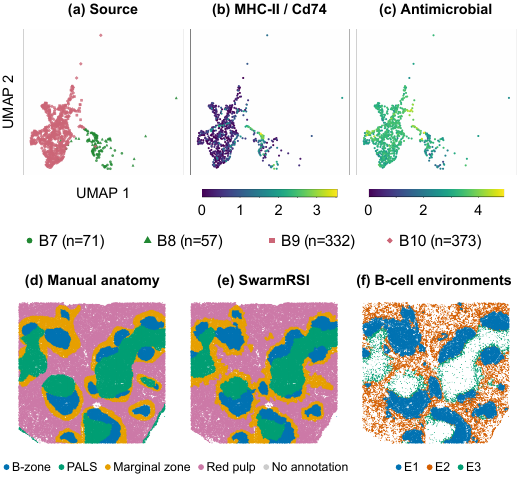}
\caption{Biological case studies. (a--c) UMAP of the reproduced \textsc{SwarmRSI} representation for 833 target enterocytes, coloured by source and by MHC-II/Cd74 and antimicrobial RNA scores (mean normalized log RNA). Green and pink denote helminth day-10 and Salmonella sources. (d--f) BALBc-1 spleen with manual annotations, \textsc{SwarmRSI} program partitions, and expert neighbourhood groups E1 (B/follicular-dendritic-cell-rich), E2 (myeloid/erythroid-rich) and E3 (T-cell-rich).}
\label{fig:enterocyte_umap_main}
\label{fig:cyto_insight_main}
\end{figure}

We examine the scientific utility of programs evolved by \textsc{SwarmRSI}. Biologists reproduced two of its programs and interpreted them with measured RNA and protein.

\Needspace{3\baselineskip}
\ParaTitle{Enterocyte: infection-associated immune expression.}
On the contrastiveVI task~\citep{weinberger2023contrastivevi}, the evolved program removes from each infected cell the mean of its healthy neighbours before reducing the residuals.
On the Haber intestinal instance~\citep{haber2017intestinal}, measured enterocyte RNA shows higher MHC-II scores in helminth day~10 than Salmonella infection, with means of 1.17 and 0.34.
\mbox{\Cref{fig:enterocyte_umap_main}a--c} shows the corresponding expression gradients on the reproduced representation.
Within the two helminth day-10 sources, MHC-II scores are locally more coherent on this representation than on RNA PCA, with Spearman's $\rho$ after covariate adjustment of 0.77 and 0.69 against 0.65 and 0.40, whereas antimicrobial scores are less locally coherent than on RNA PCA in each of the four sources.

\Needspace{3\baselineskip}
\ParaTitle{CytoCommunity: spatial context without consistent marker shifts.}
On the CytoCommunity task~\citep{hu2024cytocommunity}, the evolved program combines anatomical priors derived from phenotypes with an anatomically constrained Markov random field, raising $g$ from 0.045 to 0.102.
\Cref{fig:cyto_insight_main}d--f compares its partitions with manual anatomy and experts' subsequent grouping of B-cell neighbourhoods by local cell composition.
Across three spleen images, 20--21\% of B cells in the manually annotated marginal zone occupy myeloid/erythroid-rich neighbourhoods.
\citet{goltsev2018codex} reported that B220 and CD79b levels on splenic B cells vary with the local neighbourhood across the spleen.
Within the marginal zone, after matching these B cells on location and phenotype within each image, these and four other markers show no consistent shift between neighbourhoods across the three images, and the B220 effect changes sign.

\FloatBarrier
\section{Conclusion}
\label{sec:conclusion}

We began with two obstacles for scientific agent swarms. A failed experiment does not show whether its hypothesis or its implementation was at fault, and a program without overall gains can still hold a component that another agent needs.
\textsc{SwarmRSI} addresses both. Feedback-Guided Swarm Mutation uses other agents' feedback to decide whether to revise a hypothesis or repair its program, and Complementary Swarm Recombination carries retained hypotheses and useful components into other agents' work. Recursive Swarm Evolution returns every outcome to the shared graph, so each experiment changes the states that later agents start from.
It achieves SOTA results on OpenProblems and NatureBench and outperforms baselines with the same backbone, most final NatureBench programs build on such reuse, and biologists reproduce its programs for follow-up analyses of intestinal and splenic immunity.
A swarm thus improves its own research by evolving what its agents inherit.

\setlength{\textfloatsep}{\maintextfloatsep}
\setlength{\floatsep}{\mainfloatsep}
\setlength{\dbltextfloatsep}{\maindbltextfloatsep}
\setlength{\dblfloatsep}{\maindblfloatsep}
\captionsetup[figure]{skip=10pt}
\bibliography{custom}

\clearpage
\appendix
\raggedbottom
\setcounter{dbltopnumber}{2}
\makeatletter
\setlength{\@fptop}{0pt}
\setlength{\@dblfptop}{0pt}
\setlength{\@fpsep}{12pt}
\setlength{\@dblfpsep}{12pt}
\makeatother
\renewcommand{\topfraction}{0.95}
\renewcommand{\bottomfraction}{0.95}
\renewcommand{\textfraction}{0.05}
\renewcommand{\floatpagefraction}{0.75}
\renewcommand{\dblfloatpagefraction}{0.75}
\renewcommand{\dbltopfraction}{0.85}

\FloatBarrier
\section{Method Details}
\label{app:method_details}
This section supplements \Cref{sec:method} with details of Recursive Swarm Evolution, execution feedback, and Complementary Swarm Recombination.

\begin{algorithm}[h]
\caption{\textsc{SwarmRSI} search loop}
\label{alg:hpcoevolve}
\small
\begin{algorithmic}[1]
\REQUIRE goal $q$, starter program $p_0$, utility $u$, budget $B$
\STATE Initialize $\mathcal G_0$ with root drafts from $p_0$, and set $\mathcal F_H,\mathcal F_P\leftarrow$ initial frontiers
\WHILE{budget $B$ remains}
  \STATE Choose frontier $f$ by priority rules and \Cref{eq:frontier_score}
  \STATE Choose H-/P-mutation, rebase or crossover in $f$
  \IF{rebase}
    \STATE keep $h$ and select base $p'\in\mathcal F_P$ by need--label overlap, or the new incumbent
  \ELSIF{crossover}
    \STATE select donors $\mathcal D(p)$ from per-instance utilities $u_k$
  \ENDIF
  \STATE Planner forms experiment $(h,p)$, coder implements, evaluate $u$
  \STATE Code and scientific audits $\rightarrow E$, append $\xi=(H,P,X,E)$ to $\mathcal G$
  \STATE Emit requests: failure $\rightarrow$ deepen, mixed per-instance $\rightarrow$ broaden, new incumbent $\rightarrow$ deepen retained $h$
  \STATE Update $\mathcal F_H,\mathcal F_P$, $(N_f,G_f,C_f)$ and $U_f$
\ENDWHILE
\RETURN best valid program in $\mathcal G$
\end{algorithmic}
\end{algorithm}
\subsection{Recursive Swarm Evolution}
\label{app:selection_details}

The hypothesis frontier $\mathcal F_H$ retains active hypotheses balanced across mechanism families. The program frontier $\mathcal F_P$ retains the incumbent $p^\star$ and the best program of each realized family.
\emph{H-mutation} revises the hypothesis toward an alternative modeling or algorithmic mechanism, whereas \emph{P-mutation} keeps the hypothesis and refines an existing program lineage.

\Cref{tab:selection_config} lists the shared selection settings for DeepSeek-V4-Pro and Gemini~3.1 Pro Preview across all three benchmarks.

\begin{table}[h]
\centering
\small
\setlength{\tabcolsep}{4pt}
\begin{tabularx}{\columnwidth}{@{}Xr@{}}
\toprule
Parameter & Value\\
\midrule
Discount factor $\rho$ & 0.9\\
Exploration weight $\beta$ & 0.5\\
Request-pressure weight $\gamma$ & 0.25\\
Request-pressure saturation $K$ & 4\\
Eligible-wait threshold $W$ & 8\\
Tokens per cost unit $T$ & 1,000\\
Improvement scale $s$ & 1.0\\
Active hypotheses / program representatives & 4 / 8\\
Candidate pairs per hypothesis & 2\\
Sources per hypothesis adaptation & 1\\
Scored local continuations / local attempts & 2 / 4\\
Local stagnation patience & 1\\
Initial root drafts & 3\\
\bottomrule
\end{tabularx}
\caption{Shared H/P-selection parameters and search limits across both backbones and all three benchmarks.}
\label{tab:selection_config}
\end{table}

\ParaTitle{Improvement and computational cost.}
For an experiment initiated by frontier $a_t\in\{H,P\}$, let $u_t$ be its utility and $u_t^-$ the preceding incumbent utility.
Reward and normalized token cost are
\begin{equation}
\begin{aligned}
 r_t&=\min\!\left(1,\max\!\left(0,\frac{u_t-u_t^-}{s}\right)\right),\\
 c_t&=\max\!\left(1,\frac{\tau_t}{T}\right),
\end{aligned}
\end{equation}
where $\tau_t$ includes the experiment's planner and coder tokens.
Without a prior incumbent, reward is zero.
Failed or unscored outcomes receive zero reward and keep their cost, and infrastructure failures leave the statistics unchanged.

\ParaTitle{Selection score updates.}
Initialize $N_f=G_f=C_f=0$ for both frontiers.
After each experiment other than an infrastructure failure, update both frontiers:
\begin{equation}
\begin{aligned}
 N_f &\leftarrow \rho N_f+\mathbf{1}[f=a_t],\\
 G_f &\leftarrow \rho G_f+\mathbf{1}[f=a_t]r_t,\\
 C_f &\leftarrow \rho C_f+\mathbf{1}[f=a_t]c_t.
\end{aligned}
\end{equation}
The controller scores each sampled, feasible frontier using \Cref{eq:frontier_score}, with $\hat\mu_f=G_f/C_f$ as recent improvement per cost.
For $U_f$ unresolved requests addressable by frontier $f$, request pressure is $\psi_f=\min(U_f,K)/K$.
These requests are generated by experiments in both frontiers, coupling their allocation. Setting $\gamma=0$ removes request pressure from the index.
Discounting emphasizes recent outcomes, while request pressure saturates at $K=4$.

\ParaTitle{Frontier scheduling.}
A sole feasible frontier is selected directly.
When both are feasible, frontiers without an updated sample take priority, with H preceding P.
Subsequently, a frontier waiting for $W$ eligible choices takes priority. Otherwise, the highest index wins.
Ties favor lower discounted selection count, then H. Waiting counters advance only when both frontiers compete.
Local program refinement may precede the next frontier choice, subject to the continuation, attempt, and stagnation limits in \Cref{tab:selection_config}.
The index adapts discounted UCB to allocation between the hypothesis and program frontiers.

\ParaTitle{Ablation without H/P selection.}
The \emph{w/o H/P selection} variant in \Cref{tab:ablation} replaces this allocation rule with H/P alternation, starting with H and selecting the other frontier if the scheduled one is infeasible.
Both frontiers, H- and P-mutation, the shared graph, audits, local refinement, recombination, and total search budget remain unchanged.

\subsection{Execution and Feedback}
\label{app:execution_details}

The execution harness and search controller perform the online audits automatically.
The code audit uses execution status, artifact validity and evaluator feedback to check execution and the task contract.
The scientific audit screens implementation evidence by comparing parent and offspring source code and recorded submitted outputs.
It flags unchanged source or syntax trees, newly added functions without static name references, or identical submitted outputs when comparable output records are available.
The controller uses these source/output flags to gate hypothesis credit.
A valid scored candidate without such a flag receives hypothesis credit~\citep{potter2000cooperative}. Flagged or unscored attempts retain program-level evidence and unresolved implementation requests.
Hypothesis credit records a valid scored candidate passing the source/output checks. Improvement per cost drives H/P selection.
Failed states and states without gain remain available in the graph together with their resulting programs, outcomes, and audit evidence.

An implementation failure retains the hypothesis and emits a \emph{deepen} request for program refinement.
Mixed per-instance gains and losses emit a \emph{broaden} request for hypothesis exploration.
An improved incumbent also creates deepen requests for hypotheses previously realized on older programs.
These requests retain unresolved needs for later mutation or recombination.

\subsection{Complementary Swarm Recombination}
\label{app:adaptation_details}

\ParaTitle{Rebase selection.}
Rebase preserves hypothesis content and selects a different program from $\mathcal F_P$ as its implementation base.
Candidates are ranked by the overlap between the recorded need and mechanism labels associated with scored, unflagged programs.
A request triggered by a new incumbent directly targets that program.
The coder then realizes the hypothesis on the selected base, so that for a selected base $b_t$
\begin{equation}
 (\tilde h_t,\tilde p_t)=\bigl(h_t,\operatorname{Rebase}(h_t,b_t;s_t)\bigr).
 \label{eq:hypothesis_rebase}
\end{equation}

\ParaTitle{Crossover selection.}
\Cref{eq:instance_sources} defines candidate donors and their priorities from observed per-instance utilities.
At most two donors are selected by their largest per-instance advantage.
Let $\widehat{\mathcal D}_t$ contain the selected donors. Sources are assigned before implementation as
\begin{equation}
 \sigma_t(k)=
 \begin{cases}
 p_k^\star,&p_k^\star\in\widehat{\mathcal D}_t\text{ and }u_k(p_k^\star)>u_k(p_t),\\
 p_t,&\text{otherwise},
 \end{cases}
 \label{eq:component_sources}
\end{equation}
so the offspring uses donor components on the instances they win and retains the parent's components elsewhere.
This assignment is fixed from measured feedback before implementation. Crossover is inactive for aggregate-only feedback.
Both operators produce new candidates evaluated through the shared execution and audit loop.

\ParaTitle{State update.}
After each evaluation, the record $\xi_t$ updates the shared graph, the next acting agent assembles its state, and a mutation or recombination operation generates the next candidate,
\begin{equation}
 \begin{aligned}
 \mathcal G_{t+1}&=\operatorname{Update}(\mathcal G_t,\xi_t),\\
 s_{t+1}&=\operatorname{Assemble}(q,\mathcal G_{t+1}),\\
 (\tilde h_{t+1},\tilde p_{t+1})&=\operatorname{Generate}(q;s_{t+1}).
 \end{aligned}
 \label{eq:recursive_swarm_update}
\end{equation}
$\operatorname{Update}$ retains the record and refreshes the associated research needs, $\operatorname{Assemble}$ collects the records relevant to the next agent, and $\operatorname{Generate}$ applies an eligible operation to the selected sources.

\FloatBarrier
\section{Experimental Setup}
\label{app:implementation}

\subsection{Benchmarks}
\label{app:benchmark_details}
\paragraph{OpenProblems.}
OpenProblems follows the six-task single-cell panel used by TusoAI~\citep{turcan2026tusoai}: denoising, label prediction, batch integration, spatially variable gene identification, decomposition and visualization.
\paragraph{NatureBench.}
NatureBench~\citep{wang2026naturebench} includes 90 tasks in protein biology, cellular omics, physical modeling, molecular design, relational reasoning and biomedical modeling.
\paragraph{MLE-Bench Lite.}
MLE-Bench Lite~\citep{chan2025mlebench} comprises 22 low-complexity competitions involving model construction and competition submission.

\subsection{Baselines}
\label{app:baseline_details}
\paragraph{Benchmark-specific comparisons.}
On OpenProblems, we compare with expert-designed methods
and ChatGPT reported by TusoAI~\citep{turcan2026tusoai},
Biomni~\citep{huang2026biomni},
AIDE~\citep{jiang2025aide}, and TusoAI.
On NatureBench, we compare with
Claude Code~\citep{anthropic2026claudecode},
Codex and Gemini CLI~\citep{wang2026naturebench},
Arbor~\citep{jin2026arbor}, and AIDE.
NatureBench results without $\dagger$ in \Cref{tab:naturebench_results} are taken from the official NatureBench leaderboard~\citep{naturebenchleaderboard2026} (accessed 7 October 2026).
These entries follow NatureBench's standard protocol. Comparisons with published results are descriptive.
On MLE-Bench Lite, we compare with
MLEvolve~\citep{du2026mlevolve},
ML-Master~2.0~\citep{liu2026mlmaster2},
Arbor~\citep{jin2026arbor},
and
Famou-Agent~2.0~\citep{fmagent2025}.
Reported competition results are taken from the official
MLE-Bench leaderboard~\citep{mlebenchleaderboard2026} or the original papers~\citep{jin2026arbor,du2026mlevolve}.

\paragraph{Matched-backbone reruns.}
We rerun AIDE and TusoAI on OpenProblems, AIDE and Arbor on NatureBench, and MLEvolve and Arbor on MLE-Bench Lite using DeepSeek-V4-Pro.
We do not rerun MLEvolve on NatureBench because its harness uses MLE-Bench's competition submission and grading format~\citep{du2026mlevolve}.
AIDE serves as the program-evolution representative on NatureBench.

\subsection{Evaluation Details}
\label{app:evaluation_details}
\paragraph{Grading procedures.}
OpenProblems uses each task's official evaluator.
NatureBench results are assessed by the benchmark's validity judge
before computing Match-SOTA and Surpass-SOTA rates.
MLE-Bench Lite submissions are graded by the official grader
on the private test split.

\paragraph{Metric definitions.}
OpenProblems uses the normalized task scores reported by TusoAI~\citep{turcan2026tusoai}, with higher values indicating better performance.
Scores are scaled by 100. Avg is the unweighted mean over the six tasks, using each task's benchmark-defined normalization.
For NatureBench, the instance-level direction-adjusted improvement is
\begin{equation}
  g_i = d_i\frac{m_i-m_i^{\mathrm{SOTA}}}{|m_i^{\mathrm{SOTA}}|},
  \label{eq:evaluation_gap}
\end{equation}
where $m_i$ is the score, $m_i^{\mathrm{SOTA}}$ is its reference and $d_i\in\{-1,+1\}$ specifies the metric direction.
A task's $g$ averages its instance-level improvements.
Match-SOTA counts judge-valid tasks with $g\geq0$. Surpass-SOTA counts those with $g>0.1$.
Both rates use all 90 tasks, retaining invalid and missing results in the denominator. S and M denote Surpass-SOTA and Match-SOTA.
For the ablations, Mean $g$ (\%) is $100\times\frac{1}{90}\sum_{j=1}^{90}g_j$, where $g_j$ is task $j$'s relative improvement. All ablation tasks have valid results.
MLE-Bench Lite reports Above Median, exclusive Bronze, Silver and Gold rates, and Any-Medal, the fraction of all 22 tasks earning any medal.
Invalid or missing submissions receive no medal. All rates use the full task set.

\paragraph{Aggregation across runs.}
For \textsc{SwarmRSI} and matched-backbone reruns in Tables~\ref{tab:openproblems_results}--\ref{tab:mle_results}, we compute each benchmark metric separately for each of three seeds, then average the three values equally.
Each seed covers the same six OpenProblems tasks, 90 NatureBench tasks, or 22 MLE-Bench Lite competitions.
Task scores and domain-specific rates are averaged in the same way.
Results in Tables~\ref{tab:openproblems_results}--\ref{tab:mle_results} are reported to one decimal place after aggregation.
Published reference results retain their original reporting protocols.
Medal rates are rounded independently, so their displayed sum may differ from Any-Medal by 0.1 pp.

\paragraph{Token consumption.}
For \Cref{fig:performance_token_tradeoff}, we sum uncached input and output tokens from all LLM calls throughout each task's search.
We report the mean over all tasks in each benchmark and the three seeds, in millions of tokens per task.
Arbor's token consumption is on the same scale as the 20.19M tokens per task reported in its original paper~\citep{jin2026arbor}.

\paragraph{Significance testing.}
We compare NatureBench Match-SOTA outcomes for \textsc{SwarmRSI}, Arbor and AIDE with DeepSeek-V4-Pro, using the 90 tasks as paired units.
Each task contributes its mean binary Match-SOTA outcome across the three runs.
As \Cref{app:statistical_reliability} explains, each method's task-level Match-SOTA outcomes are identical across its three seeds.
We resample the 90 paired task means with replacement 10,000 times and recompute the rate difference in each replicate~\citep{koehn2004statistical}.
The 2.5th and 97.5th percentiles give the 95\% paired bootstrap confidence interval.
The Match-SOTA improvement over Arbor is 7.8 pp (95\% CI [2.2, 13.3]). The improvement over AIDE is 16.7 pp (95\% CI [8.9, 24.4]).
Exact McNemar tests use the discordant task pairs: \textsc{SwarmRSI}-only/baseline-only successes are 7/0 against Arbor and 15/0 against AIDE.
The resulting $p$-values are $0.015625$ and $0.0000610$, respectively.
Differences are computed before rounding the displayed rates. Other benchmark and leaderboard comparisons are reported descriptively.

\subsection{Implementation Details}
\label{app:implementation_details}
\paragraph{Backbones.}
\textsc{SwarmRSI} and the rerun baselines use DeepSeek-V4-Pro~\citep{deepseek2026v4}.
Additional \textsc{SwarmRSI} runs use Gemini~3.1 Pro Preview~\citep{google2026gemini31pro}.

\paragraph{Time budgets.}
We set per-task time limits to 8 hours on
OpenProblems~\citep{turcan2026tusoai}, 4 hours on
NatureBench~\citep{wang2026naturebench}, and 12 hours on
MLE-Bench Lite~\citep{du2026mlevolve}.
Within each benchmark, \textsc{SwarmRSI} and rerun baselines
share the evaluation protocol and time limit.

\paragraph{Hardware configurations.}
\textsc{SwarmRSI} and rerun baselines use the same hardware within each benchmark. OpenProblems runs use 4 CPU threads and 50 GB of memory, each of the 87 NatureBench tasks that require a GPU uses one GPU, and MLE-Bench Lite runs use 21 vCPUs, 234 GB of RAM and one GPU.
These follow the published protocols for OpenProblems~\citep{turcan2026tusoai}, NatureBench~\citep{wang2026naturebench}, and MLE-Bench Lite~\citep{du2026mlevolve}.

\subsection{Ablation Settings and Aggregation}
\label{app:ablation_details}
The variants in \Cref{tab:ablation} use the same planner, coder, evaluator and total search budget.
The w/o shared graph variant preserves hypotheses, executable programs, and experimental feedback while disabling their graph-based organization and reuse.
The w/o H-mutation variant only repairs programs under existing hypotheses ($\tilde h_t=h_t$), while w/o P-mutation only revises hypotheses.
The w/o recombination variant disables both rebase and crossover in Complementary Swarm Recombination, retaining within-trajectory evolution.
The w/o H/P selection variant, described in \Cref{app:selection_details}, retains both frontiers and replaces adaptive H/P selection with fixed H/P alternation.

For seed $s$, let $n_s^{M}$ and $n_s^{S}$ count judge-valid tasks with $g\geq0$ and $g>0.1$, respectively.
The reported rates are $\frac{100}{3}\sum_{s=1}^{3}n_s^{M}/90$ and $\frac{100}{3}\sum_{s=1}^{3}n_s^{S}/90$.
Mean $g$ is $\frac{100}{270}\sum_{s=1}^{3}\sum_{t=1}^{90}g_{s,t}$. All ablation tasks have valid results.
Thus, each seed contributes equally, and values are rounded to one decimal place only after averaging.

\subsection{Performance and Token Cost}
\label{app:performance_cost}
\begin{figure}[!htbp]
\centering
\includegraphics[width=\columnwidth]{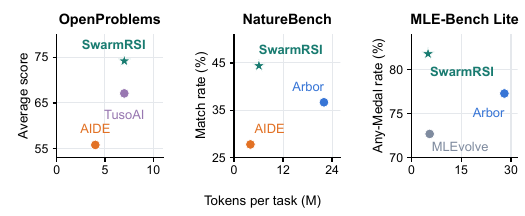}
\caption{Performance against mean tokens per task for \textsc{SwarmRSI} and rerun baselines with DeepSeek-V4-Pro. OpenProblems reports normalized average scores scaled by 100, and NatureBench and MLE-Bench Lite report Match-SOTA and Any-Medal rates (\%).}
\label{fig:performance_token_tradeoff}
\end{figure}

\Cref{fig:performance_token_tradeoff} plots benchmark performance against mean tokens per task for \textsc{SwarmRSI} and the rerun baselines with DeepSeek-V4-Pro.

\subsection{Statistical Reliability}
\label{app:statistical_reliability}
\Cref{tab:ablation_ci} reports, for each ablation in \Cref{tab:ablation}, the drop $\Delta g$ in mean $g$ from \textsc{SwarmRSI} with a task-paired bootstrap 95\% confidence interval over the 90 NatureBench tasks. All intervals exclude zero.

\begin{table}[!htbp]
\centering
\small
\begin{tabular}{@{}lcc@{}}
\toprule
Variant & $\Delta g$ & 95\% CI\\
\midrule
w/o H-mutation & 5.6 & [1.2, 10.2]\\
w/o P-mutation & 7.4 & [3.0, 11.9]\\
w/o recombination & 5.2 & [2.5, 7.9]\\
w/o shared graph & 4.7 & [0.3, 9.3]\\
w/o H/P selection & 8.2 & [4.2, 12.4]\\
\bottomrule
\end{tabular}
\caption{Drop in mean $g$ (scaled by 100) from \textsc{SwarmRSI} for each ablation, with task-paired bootstrap 95\% CIs.}
\label{tab:ablation_ci}
\end{table}

We evaluate all methods marked with $\dagger$ in \Cref{tab:openproblems_results} using three independent random seeds.
We report the mean performance and compute 95\% confidence intervals using the Student's $t$-distribution:
\[
\mathrm{CI}_{95\%}
=
\bar{x} \pm t_{0.975,2}\frac{s}{\sqrt{3}},
\]
where $\bar{x}$ and $s$ denote the sample mean and sample standard deviation across the three seeds, respectively.
\Cref{tab:perseed} reports the per-seed OpenProblems scores, and \Cref{tab:openproblems_ci} reports their 95\% CI half-widths.
The Avg CI is computed from the three per-seed averages before rounding.

\paragraph{Thresholded outcomes.}
In the main experiments of Tables~\ref{tab:openproblems_results}--\ref{tab:mle_results} and \Cref{tab:ablation}, task-level $g$ scores varied across three independent seeds, with per-task ranges of $0.0123$--$0.0781$.
However, no task crossed the Match-SOTA ($g=0$) or Surpass-SOTA ($g=0.1$) threshold, and no MLE-Bench Lite competition crossed an Above-Median or medal boundary.
Consequently, the thresholded success rates remained identical across seeds, despite variation in the underlying continuous scores.

\begin{table*}[t]
\centering
\footnotesize
\captionsetup{font=footnotesize,skip=3pt}
\setlength{\tabcolsep}{4pt}
\renewcommand{\arraystretch}{1.08}
\begin{tabular}{llccccccc}
\toprule
Method & Seed & Denoise & Label & Batch & SVG & Decomp & Visual & Avg \\
\midrule
\multirow{3}{*}{\textsc{SwarmRSI} (DeepSeek)}
 & 1 & 34.1 & 74.6 & 83.1 & 84.7 & 85.6 & 85.2 & 74.55 \\
 & 2 & 33.5 & 73.5 & 82.8 & 83.8 & 86.6 & 84.6 & 74.13 \\
 & 3 & 33.3 & 74.4 & 81.6 & 83.1 & 85.8 & 84.7 & 73.82 \\
\midrule
\multirow{3}{*}{\textsc{SwarmRSI} (Gemini)}
 & 1 & 40.2 & 81.5 & 79.0 & 86.6 & 82.7 & 85.5 & 75.92 \\
 & 2 & 40.2 & 80.7 & 78.1 & 88.8 & 84.0 & 86.1 & 76.32 \\
 & 3 & 40.3 & 80.0 & 79.4 & 88.1 & 83.6 & 87.5 & 76.48 \\
\midrule
\multirow{3}{*}{AIDE (DeepSeek)}
 & 1 & 27.3 & 70.5 & 78.6 & 72.2 & 46.6 & 40.4 & 55.93 \\
 & 2 & 28.5 & 70.2 & 77.5 & 74.5 & 45.4 & 39.6 & 55.95 \\
 & 3 & 26.4 & 71.2 & 77.9 & 71.6 & 46.4 & 39.5 & 55.50 \\
\midrule
\multirow{3}{*}{TusoAI (DeepSeek)}
 & 1 & 29.3 & 57.5 & 79.5 & 79.0 & 84.6 & 69.9 & 66.63 \\
 & 2 & 32.9 & 59.1 & 78.8 & 80.1 & 83.9 & 69.7 & 67.42 \\
 & 3 & 32.2 & 59.5 & 80.1 & 77.9 & 83.9 & 69.9 & 67.25 \\
\bottomrule
\end{tabular}
\caption{Per-seed OpenProblems scores for all $\dagger$ entries in \Cref{tab:openproblems_results}. Scores are normalized and scaled by 100. Avg is the six-task mean, shown to two decimal places. DeepSeek and Gemini denote DeepSeek-V4-Pro and Gemini 3.1 Pro Preview, respectively.}
\label{tab:perseed}
\end{table*}

\begin{table*}[t]
\centering
\footnotesize
\captionsetup{font=footnotesize,skip=3pt}
\setlength{\tabcolsep}{5pt}
\renewcommand{\arraystretch}{1.08}
\begin{tabular}{lccccccc}
\toprule
Method & Denoise & Label & Batch & SVG & Decomp & Visual & Avg \\
\midrule
\textsc{SwarmRSI} (DeepSeek) & 1.03 & 1.46 & 1.97 & 1.99 & 1.31 & 0.80 & 0.91 \\
\textsc{SwarmRSI} (Gemini) & 0.14 & 1.86 & 1.65 & 2.79 & 1.65 & 2.55 & 0.72 \\
AIDE (DeepSeek) & 2.62 & 1.27 & 1.38 & 3.80 & 1.60 & 1.23 & 0.63 \\
TusoAI (DeepSeek) & 4.74 & 2.63 & 1.62 & 2.73 & 1.00 & 0.29 & 1.03 \\
\bottomrule
\end{tabular}
\caption{95\% CI half-widths for the OpenProblems means over three seeds in \Cref{tab:openproblems_results}, computed using the Student's $t$-distribution from \Cref{tab:perseed}. Intervals are the unrounded means $\pm$ these half-widths, on the score scale of \Cref{tab:openproblems_results}.}
\label{tab:openproblems_ci}
\end{table*}

\FloatBarrier
\section{Additional Trajectory Analyses}
\label{app:trajectory_details}
\Cref{app:trace_comparison} gives trajectory construction and operator-level evidence for \Cref{subsec:analysis}, and \Cref{app:search_diagnostics} gives aggregate search statistics.
\subsection{Comparative Trajectories}
\label{app:trace_comparison}

We analyze one recorded run per method and task.
The two examples in \Cref{fig:trajectories}c,d illustrate successive hypothesis mutations on 00917, and crossover followed by rebase on 00639.
The plotted runs use DeepSeek-V4-Pro. Candidate scores and operator sequences accompany the plotting code.

\ParaTitle{Trajectory construction.}
For each method, a scored evaluation is a recorded finite score accepted by its evaluator interface. AIDE nodes marked buggy or worst are excluded from this axis.
We plot $B_k=\max_{i\leq k}g_i$ over these records, with stepwise interpolation and no extrapolation beyond the last observation.
Unscored records are retained in the source data and counted separately below.
Arbor records evaluator calls, whereas AIDE records candidate nodes. The plot uses scored evaluations for both methods.

\begin{table}[h]
\centering\small
\begin{tabular}{llrr}
\toprule
Task & Method & Scored/records & Best $g$\\
\midrule
00917 & \textsc{SwarmRSI} & 18/18 & 0.1807\\
 & Arbor & 3/4 & -0.2007\\
 & AIDE & 49/50 & -0.4742\\
00639 & \textsc{SwarmRSI} & 19/21 & 0.0644\\
 & Arbor & 2/3 & 0.0434\\
 & AIDE & 8/28 & 0.0552\\
\bottomrule
\end{tabular}
\caption{Selected runs with their scored evaluations, recorded events and best observed $g$.}
\end{table}

\ParaTitle{Analysis of search operators.}
Operator markers use the frontier episodes associated with each candidate's action ID.
In 00917, candidate 3 is a program mutation of candidate 1. Its $g=-0.9903$ does not improve the incumbent ($g=-0.0501$).
In 00917, candidate 6 retains the recurrent shell-sequence program at $g=-0.0162$. Candidates 7--15 do not improve this incumbent.
Hypothesis-mutation candidates 16 and 17 descend from candidates 6 and 16, respectively, replacing the readout with normalized allocation and then a saturating absorption--yield readout.
Both modify the executed readout and are reevaluated, reaching $g=0.0928$ (Match-SOTA) and $0.1807$ (Surpass-SOTA).
In 00639, candidate 8 (zero-based index, scored evaluation 7) uses candidate 4 for esol, freesolv, and lipophilicity, candidate 1 for fluorescence and stability, and candidate 0 for the two reaction instances.
Its code reuses separate descriptor, neural-sequence, and hashed-reaction branches from the source programs.
Candidate 19 includes message-passing and ranking-loss functions, but its entry-point routing does not execute them.
Candidate 20, recorded as rebase, activates those paths for lipophilicity and fluorescence, respectively.
Their instance improvements change from $0.1002581$ to $0.1337743$ and from $-0.0201618$ to $-0.0056382$. The other five instance scores remain unchanged.
The mean consequently rises from $0.0575466$ to $0.0644094$.
The activated functions were already present in candidate 19 and absent from retrieved source 16. The step repairs their entry-point routing.

\ParaTitle{A retained hypothesis on a new program base.}
As shown in \Cref{tab:rebase_process}, task 00795 provides a complete sequence linking an implementation request, a changed program base, executable transfer and per-instance evaluation.
Candidate 4 proposes using a source program's molecular-function predictor on one instance while retaining the parent's other four predictors.
Its submitted outputs are identical to candidate 3's, so the controller retains the hypothesis with an unresolved implementation request.
Candidate 8 subsequently improves the program through hypothesis mutation.
Rebase candidate 9 returns to the retained hypothesis on this newer base, with source program 1 supplying the implementation context.
The code disables the ranking-loss term for the target predictor and adopts the source's 30-threshold public-validation checkpoint rule. The target branch executes these settings and writes its predictions.
The target Fmax rises from $0.5032$ to $0.5087$, while the other four instance scores remain unchanged.
This raises $g$ from $-0.06388$ to $-0.06186$.

\begin{table}[h]
\centering\small
\setlength{\tabcolsep}{3pt}
\begin{tabularx}{\columnwidth}{@{}lXr@{}}
\toprule
Candidate & Recorded change & $g$\\
\midrule
4 & Retained hypothesis. Outputs unchanged from 3 & $-0.08323$\\
8 & Hypothesis mutation produces a new program base & $-0.06388$\\
9 & Rebase: target predictor uses source training and checkpoint rules & $-0.06186$\\
11 & Crossover: source 5 supplies a second instance predictor & $-0.05573$\\
15 & Rebase: target predictor implemented without fallback & $-0.05408$\\
\bottomrule
\end{tabularx}
\caption{Recorded process on 00795. Candidate indices are zero-based. Each $g$ comes from a valid evaluation of the resulting program.}
\label{tab:rebase_process}
\end{table}

Crossover candidate 11 then adds source 5's predictor on a second instance, reproducing its Fmax of $0.3773$ and preserving candidate 9's other four scores.
Its execution log records another fit and prediction pass for the first transferred predictor, retaining its Fmax of $0.5087$.
Source 5 had a lower aggregate score than its incumbent, yet its component remains useful in this new program.
Candidates 13 and 14 attempt another transfer but retain candidate 11's submitted outputs. Their implementation requests remain open.
Candidate 15 realizes the retained target predictor without falling back to the incumbent's outputs, increasing its Fmax to $0.5133$ while preserving the other four scores.
The sequence shows how an untested hypothesis survives its first implementation, is tested on a later program base, and contributes to subsequent component reuse and refinement.

\ParaTitle{Feedback-linked updates and component reuse.}
\label{app:coevolution_progress}
We analyze one complete seed (90 runs) with DeepSeek-V4-Pro and recorded request identifiers.
These runs cover all 90 NatureBench tasks and comprise 1,948 search attempts and 426 new-best candidates.
Program feedback prompted hypothesis mutations, while unresolved implementation requests for retained hypotheses prompted rebase updates on the program frontier.
We matched these updates to their originating requests using recorded request and episode identifiers.
These updates comprised 296 attempts (15.2\%) and produced 84 new-best candidates (19.7\%), a 28.4\% new-best rate versus the overall 21.9\%. Including crossover, the union of these events comprised 731 attempts (37.5\%) and produced 173 new-best candidates (40.6\%).

\ParaTitle{Cumulative progress across search stages.}
To examine whether progress builds on inherited research states, we trace each run's final best program through its recorded code-parent and intervention-source edges.
In about 77\% of runs, this lineage contains at least one rebase, crossover, or update triggered by feedback.
\Cref{tab:stage_reuse} further divides each run's attempts into early, middle and late thirds by step order.
Reuse and update triggered by feedbacks achieve higher new-best rates than other updates in every stage, and their share of new-best candidates increases slightly over the course of search.
The overall new-best rate declines as search proceeds, consistent with diminishing returns as the incumbent improves.

\begin{table}[h]
\centering
\small
\setlength{\tabcolsep}{3pt}
\begin{tabular}{@{}lccc@{}}
\toprule
Stage & Reuse hit & Other hit & Reuse share\\
\midrule
Early & 32.0 & 25.8 & 40.5\\
Middle & 18.2 & 16.8 & 41.1\\
Late & 18.1 & 15.0 & 42.6\\
\bottomrule
\end{tabular}
\caption{New-best rates (\%) of reuse and update triggered by feedbacks (rebase, crossover and mutations triggered by feedback) and of other updates across search stages on NatureBench. Reuse share is the percentage of new-best candidates from these updates.}
\label{tab:stage_reuse}
\end{table}

\subsection{Cross-Task Transfer}
\label{app:cross_task}
\paragraph{Cross-task research state inheritance.}
After each generation, the accumulated research state is preserved for subsequent tasks.
This state includes hypotheses, program components, experimental evidence, and unresolved implementation needs.
On a new task, prior records are retrieved by matching their mechanism labels to the task description and initialize the agent's research state, while candidate programs are adapted and evaluated under the new task's constraints.

\paragraph{Experimental setup and results.}
We run \textsc{SwarmRSI} with DeepSeek-V4-Pro on six NatureBench source tasks in three generations of two tasks each, and each generation starts from the cross-task state exported by the previous one.
We then search 12 held-out NatureBench tasks, fixed in advance and disjoint from the source tasks, from an empty state and from the state after each generation, with three seeds and the same budget per task.
As shown in \Cref{tab:cross_task}, mean $g$ rises from 12.01 under cold start to 16.15, 20.16 and 21.89 after one, two and three generations, and Gen~3 improves on cold start by 9.88 points.
Match-SOTA rises from 41.7\% to 86.1\% and Surpass-SOTA from 33.3\% to 58.3\%, while tokens per task fall from 12.17M to 10.94M.
Gains are largest over the first two generations, and Gen~3 matches Gen~2 on Match-SOTA and improves mainly through Surpass-SOTA.
Unlike the main experiments described in \Cref{app:statistical_reliability}, some held-out tasks cross the Match-SOTA or Surpass-SOTA threshold in only some seeds, so the rates in \Cref{tab:cross_task} count the 36 task and seed pairs, for example 31 of 36 for Match-SOTA after Gen~3.

\begin{table}[h]
\centering
\scriptsize
\setlength{\tabcolsep}{1.8pt}
\begin{tabular*}{\columnwidth}{@{\extracolsep{\fill}}lccccc@{}}
\toprule
Variant & Match & Surpass & Mean $g$ & $\Delta g$ [95\% CI] & Tokens\\
\midrule
Cold start & 41.7 & 33.3 & 12.01 & -- & 12.17\\
Gen 1 & 58.3 & 41.7 & 16.15 & 4.14 [1.82, 6.49] & 11.86\\
Gen 2 & 86.1 & 44.4 & 20.16 & 8.15 [5.67, 10.59] & 11.52\\
Gen 3 & \textbf{86.1} & \textbf{58.3} & \textbf{21.89} & 9.88 [6.88, 13.06] & \textbf{10.94}\\
\bottomrule
\end{tabular*}
\caption{Search on 12 held-out NatureBench tasks with DeepSeek-V4-Pro, starting from an empty state (cold start) or from the state inherited after each generation, averaged over three seeds. Match and Surpass are Match-SOTA and Surpass-SOTA rates (\%), mean $g$ is scaled by 100, and $\Delta g$ is the gain over cold start with a paired task-cluster bootstrap 95\% CI from 10,000 resamples. Tokens are means per task in millions.}
\label{tab:cross_task}
\end{table}

\subsection{Aggregate Search Behavior}
\label{app:search_diagnostics}

\ParaTitle{Program mutation.}
\Cref{fig:search_diagnostics}a compares \textsc{SwarmRSI} with AIDE and Arbor across all 90 NatureBench tasks.
Stagnation, the share of scored candidates that do not improve the preceding best valid result, is 43\% for \textsc{SwarmRSI}, compared with 63\% for AIDE and 70\% for Arbor.
Failure recovery is the share of failed implementations whose subsequent refinement yields a new best valid program, with failed implementations as the denominator.

\ParaTitle{Hypothesis mutation.}
\Cref{fig:search_diagnostics}b reports the proportions of proposed hypotheses that are implemented, evaluated with a valid numeric score, and associated with an improvement over the preceding best valid result.
For each method, all three proportions use the total number of proposed hypotheses in its search records as the denominator.

\FloatBarrier
\section{Human Failure Analysis}
\label{app:failure_analysis}

\paragraph{Sample and unit of analysis.}
We reviewed 35 failed \textsc{SwarmRSI} task trajectories across NatureBench and MLE-Bench Lite.
The annotation unit was an individual attempt, assessed against the best preceding valid result using each benchmark's metric direction.
The final failure set comprised 644 attempts across these trajectories, after applying the exclusions below.
OpenProblems reports continuous normalized scores without a binary success criterion, so no OpenProblems trajectories enter the failure set.
The 35 trajectories comprise 32 NatureBench tasks below Match-SOTA and 3 MLE-Bench Lite competitions without a medal.

\paragraph{Evaluation criteria.}
Annotators examined the proposed intervention, implementation and execution records, evaluator outputs, and preceding scores.
\emph{Implementation or execution failure} denotes an intervention that could not be reliably tested because of code or interface errors, execution interruptions, or proposal--implementation mismatches.
\emph{No performance gain} denotes a faithfully implemented and executed intervention with a valid evaluation that did not improve the preceding best valid result.
Equal scores were treated as no gain.
Recovered intermediate errors were judged by the final evaluated outcome.
Successful verification or restoration of an existing result, and repetition of an attained metric ceiling, were excluded from the failure counts.
Attempts with insufficient evidence to distinguish the two failure classes were also excluded.

\paragraph{Human review and agreement.}
Two annotators with PhDs in bioinformatics engineering reviewed the evidence using these criteria, with inter-rater agreement of Cohen's $\kappa=0.92$.

\paragraph{Representative cases.}
On task 00838, a proposed amino-acid substitution was overwritten by downstream rules, leaving the evaluated output unchanged This was an implementation failure despite a valid score.
In contrast, on task 02316, a negative-binomial mixture with a Potts spatial prior ran successfully but collapsed all locations into one cluster, yielding ARI zero. This was a no-gain outcome.

\FloatBarrier
\section{Biological Case Studies}
\label{app:biological_cases}
\Cref{app:case01955,app:case02124} detail the Enterocyte and CytoCommunity case studies from \Cref{subsec:case_study} with their data, programs and expert analyses.
\subsection{Enterocyte: Infection-Associated Expression Variation}
\label{app:case01955}

\paragraph{Task, data and selected program.}
NatureBench task s41592-023-01955-3 is associated with contrastiveVI~\citep{weinberger2023contrastivevi} and returns a low-dimensional target-cell representation.
The program selected by \textsc{SwarmRSI} comes from step~4 of five recorded steps.
Its recorded task-level $g=1.838461$ averages the Haber, McFarland and Norman instances.
The analysis here concerns only the Haber intestinal instance~\citep{haber2017intestinal}, GEO GSE92332.
The public matrix contains 9,842 cells and 15,215 genes: 3,240 healthy background cells, 2,711 helminth day-10 and 1,770 Salmonella target cells, and 2,121 helminth day-3 cells outside the target submission.
The published Enterocyte annotation selects 1,458 cells, including 833 targets (128 helminth day-10 and 705 Salmonella).

\paragraph{Evolved program and expert reproduction.}
For Haber, the program normalizes total counts to 10,000 and applies log1p, selects 2,000 genes by variance across target and background cells, and standardizes using background means and standard deviations, clipping at $\pm10$.
It identifies 20 healthy neighbours per target in a 50-dimensional background PCA space and subtracts their mean expression.
A singular-value decomposition yields a 20-dimensional residual space. A 15-neighbour graph then smooths the first ten residual coordinates for two iterations, including a self weight.
Experts reproduced the program. Local-coherence analysis uses its ten-dimensional outputs.
The embedding is constructed from RNA, and infection labels parsed from public cell identifiers are used for diagnostic silhouette reporting.

\begin{figure*}[p]\centering
\includegraphics[width=\textwidth]{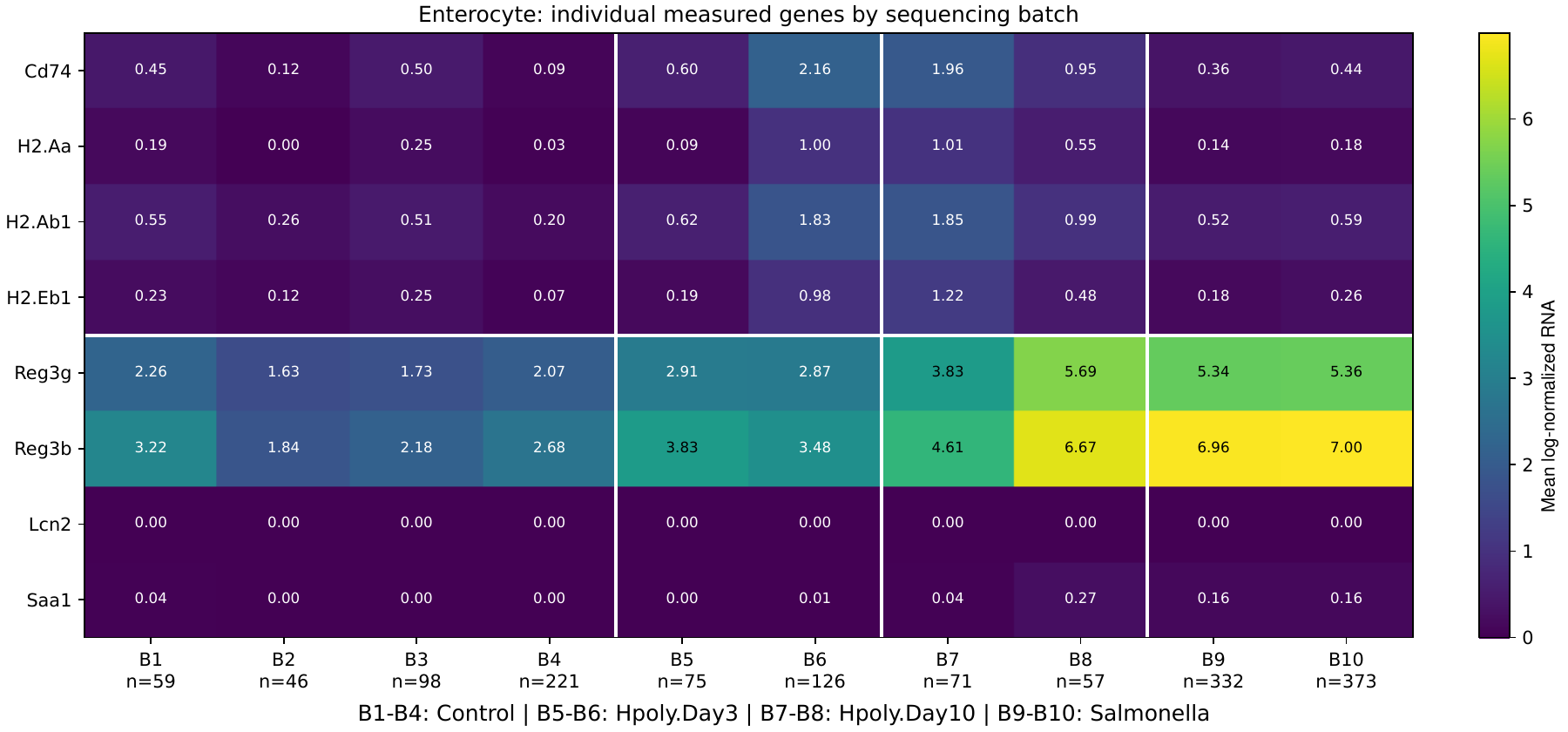}
\caption{Source means of individual measured genes in enterocytes. Hpoly.Day3/Day10 denote helminth day~3/day~10 sources. The antimicrobial set is primarily Reg3g/Reg3b. Zero Lcn2 denotes non-detection in this matrix. Values are mean normalized log RNA.}
\label{fig:enterocyte_genes}
\end{figure*}

\paragraph{Measured programs and source units.}
Each program score is the arithmetic mean of its genes' $\log(1+10^4x/\sum x)$ expression.
The MHC-II-associated set is Cd74, H2-Aa, H2-Ab1 and H2-Eb1 (the input matrix uses dots in the H2 gene names).
The interferon set is Stat1, Isg15, Ifit1, Ifit3, Oas1a and Irf7. All six genes are present in the matrix.
The antimicrobial set contains Reg3g, Reg3b, Lcn2 and Saa1. Its signal is dominated by Reg3g/Reg3b, with Lcn2 undetected and Saa1 low in these enterocytes.
B1--B4 are control sources, B5/B6 helminth day~3, B7/B8 helminth day~10, and B9/B10 Salmonella.
Source labels denote sequencing batches.
\Cref{fig:enterocyte_genes} shows that the four MHC-II genes each have higher source means in B7/B8 than in B9/B10.
The antimicrobial program means are 2.120 and 3.158 in B7/B8 versus 3.114 and 3.130 in B9/B10, showing a source-dependent contrast.
\Cref{tab:enterocyte_source_programs} reports MHC-II and interferon means for all ten sources.
Pooling enterocytes within each condition gives MHC-II means of 1.168 in helminth day~10, 0.335 in Salmonella and 0.202 in controls.
The corresponding interferon means are 0.356, 0.438 and 0.150.
These condition means weight each source by its enterocyte count.

\begin{table}[H]
\centering\small
\begin{tabular}{lrrr}\toprule
Source & Cells & MHC-II & Interferon\\\midrule
B1 & 59 & 0.354 & 0.115\\
B2 & 46 & 0.126 & 0.166\\
B3 & 98 & 0.375 & 0.136\\
B4 & 221 & 0.100 & 0.162\\
B5 & 75 & 0.377 & 0.391\\
B6 & 126 & 1.495 & 0.636\\
B7 & 71 & 1.509 & 0.395\\
B8 & 57 & 0.744 & 0.306\\
B9 & 332 & 0.299 & 0.444\\
B10 & 373 & 0.367 & 0.432\\\bottomrule
\end{tabular}
\caption{Measured enterocyte RNA program means by sequencing source. Scores average normalized log RNA across the four MHC-II or six interferon genes. B1--B4 are controls, B5/B6 helminth day~3, B7/B8 helminth day~10 and B9/B10 Salmonella.}
\label{tab:enterocyte_source_programs}
\end{table}

\begin{figure*}[p]\centering
\includegraphics[width=\textwidth]{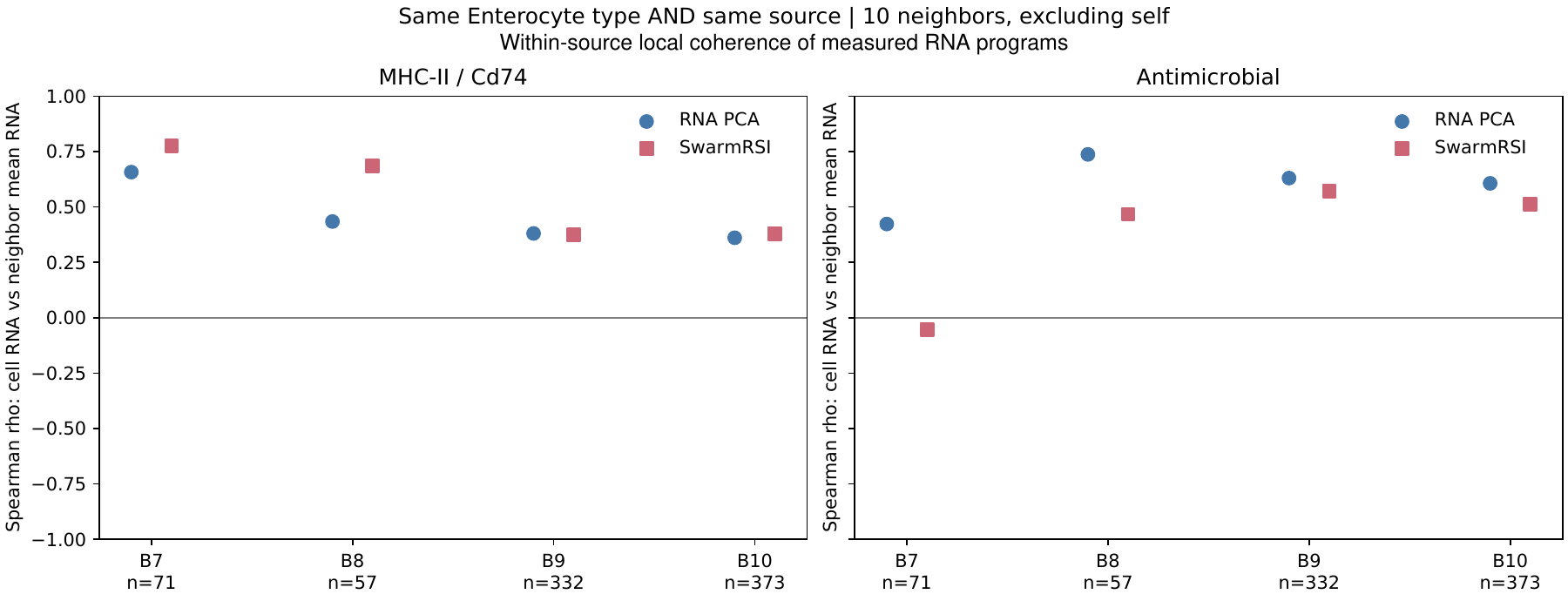}
\caption{\textbf{Within-source expression coherence in enterocytes.} Within each source, Spearman correlation compares cells' measured RNA program scores with their ten nearest neighbours' mean scores, excluding each focal cell. Neighbours are defined in RNA PCA (blue) or the reproduced \textsc{SwarmRSI} representation (pink), and the correlations are not adjusted for covariates.}
\label{fig:enterocyte_coherence}
\end{figure*}

\paragraph{Representation follow-up and sensitivity checks.}
Measured RNA and the reproduced representation are matched by cell identifier.
Within each of B7--B10, the analysis selects ten neighbours among cells of the same Enterocyte type and source, excludes the focal cell, and correlates its program score with the neighbours' mean score using Spearman's $\rho$.
\Cref{fig:enterocyte_coherence} reports these unadjusted correlations.
Sensitivity checks use five and fifteen neighbours and the representation saved from the original \textsc{SwarmRSI} run.
For covariate sensitivity, each program score is regressed within source on an intercept, log1p total UMI, log1p detected genes and the mean absorptive-marker score (Alpi, Apoa1, Fabp1, Fabp2, Sis).
\Cref{tab:enterocyte_adjusted} reports the same correlation using the residual score and the mean residual of the fixed ten neighbours.
Local coherence quantifies measured RNA structure retained in each RNA-derived representation.

\begin{table*}[!tp]
\centering\small
\begin{tabular}{llrrrr}\toprule
Source & Program & Cells & RNA PCA & Reproduced output & Recorded output\\\midrule
B7 & MHC-II & 71 & 0.650 & 0.768 & 0.766\\
B8 & MHC-II & 57 & 0.396 & 0.685 & 0.686\\
B9 & MHC-II & 332 & 0.331 & 0.326 & 0.313\\
B10 & MHC-II & 373 & 0.356 & 0.351 & 0.347\\
B7 & Antimicrobial & 71 & 0.311 & $-0.045$ & $-0.056$\\
B8 & Antimicrobial & 57 & 0.689 & 0.593 & 0.612\\
B9 & Antimicrobial & 332 & 0.370 & 0.324 & 0.331\\
B10 & Antimicrobial & 373 & 0.509 & 0.296 & 0.286\\\bottomrule
\end{tabular}
\caption{Within-source local coherence after covariate adjustment. Entries are Spearman correlations between a cell's residual RNA score and its ten neighbours' mean residual score. Reproduced/recorded outputs are the expert reproduction and the representation saved from the original \textsc{SwarmRSI} run. Counts are cells per source.}
\label{tab:enterocyte_adjusted}
\end{table*}

\paragraph{Additional visualizations.}
The following figures report within-source coherence and gene-level source means. \Cref{fig:enterocyte_umap_main}a--c shows RNA scores on the reproduced representation.

\FloatBarrier
\twocolumn[{
\subsection{CytoCommunity: Spatial Context and Protein Follow-up}
\label{app:case02124}

\begin{minipage}{\textwidth}
\centering
\includegraphics[width=\linewidth]{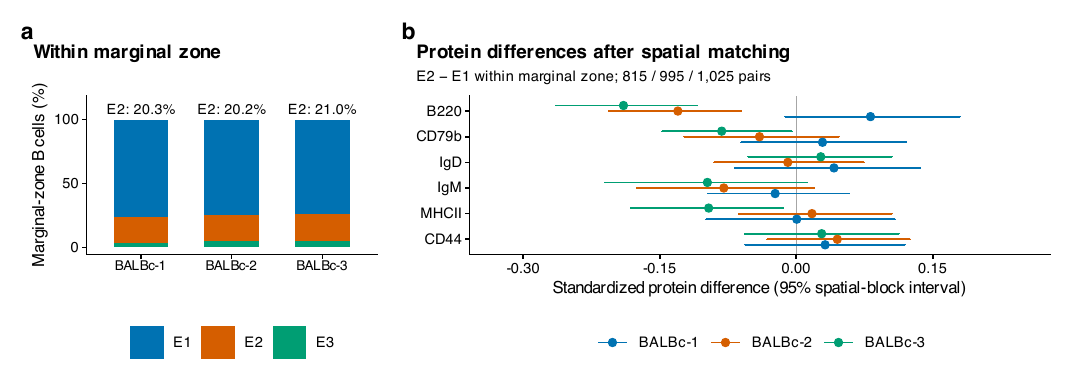}
\captionof{figure}{\textbf{Expert follow-up of B-cell neighbourhoods across three spleen images.} E1--E3 are post-hoc groups defined by local cell composition. (a) Their proportions among B cells in the manually annotated marginal zone. (b) All six primary protein contrasts after within-image spatial/phenotype matching, E2 minus E1 in that zone (815, 995 and 1,025 pairs). Points are standardized transformed-protein differences. Intervals are 95\% within-image spatial-block bootstrap intervals.}
\label{fig:cyto_quantitative}
\end{minipage}
\vspace{10pt}
}]
\paragraph{Dataset and evolved program.}
NatureBench task s41592-023-02124-2 uses three healthy mouse spleen CODEX images associated with CytoCommunity~\citep{hu2024cytocommunity}: 82,382, 81,925 and 80,973 cells, respectively, with 27 supplied cell phenotypes.
\textsc{SwarmRSI} receives coordinates and phenotype labels. The expert follow-up separately examines measured protein intensities.
The program selected by \textsc{SwarmRSI} comes from step 12 of a 15-step search and raises $g$ from 0.04507 to 0.10215.
It uses phenotype-derived anatomical priors, 50-neighbour composition and interface features, Gaussian EM, an anatomically constrained spatial MRF and cleanup of components smaller than 150 cells.
Recorded macro-F1 and AMI are 0.639247 and 0.583341. Macro-F1 includes the unannotated reference category, whose F1 is zero.
Marginal-zone F1 is 0.600, 0.593 and 0.635.

\paragraph{Post-hoc neighbourhood descriptions.}
The analysis includes 35,168, 33,544 and 32,451 conventionally annotated B cells (101,163 total).
Experts fit three exploratory neighbourhood-composition prototypes in the first image and then hold them fixed for the other two, giving B/FDC-rich (E1), myeloid/erythroid-rich (E2) and T-cell-rich (E3) groups.
E1--E3 describe local cell composition around B cells in this post-hoc analysis. They are distinct from the program's anatomical partitions.
The primary comparison uses E2 versus E1 within the \emph{manual} marginal zone, independently of whether the \textsc{SwarmRSI} program assigned that cell to the marginal zone.
Coordinates are analyzed in dataset units.

\paragraph{Expert protein and geometry checks.}
The analysis uses processed protein measurements for all 245,280 cells.
Within each image and manual marginal zone, matching is exact on original phenotype cluster and imaging tile, then uses five geometry variables with a 0.5-SD per-variable caliper, joint distance at most 1 and no replacement.
There are 815, 995 and 1,025 matched E2--E1 pairs, retaining 42.5\%, 46.2\% and 41.1\% of E2 cells, respectively.
Maximum absolute geometry standardized mean differences are 0.080, 0.063 and 0.083. Protein contrasts use this matched overlap subset.
All six preselected proteins (CD79b, B220, IgD, IgM, MHCII, CD44) are shown in \Cref{fig:cyto_quantitative}b.
Measurements are transformed as $\operatorname{arcsinh}(x/150)$, retaining negative values. Effects are differences of transformed means divided by the within-image contrast-population standard deviation.
Intervals use 1,000 bootstrap resamples of 1,000-coordinate-unit spatial blocks within each image.
B220 effects are $+0.082$, $-0.130$ and $-0.190$. Stricter 0.25-SD matching also yields inconsistent directions.
In spatially blocked validation, adding neighbourhood labels to a geometry/phenotype protein predictor produces $\Delta R^2$ from $-0.00098$ to $+0.00967$ across 18 image--protein combinations.
Neighbourhood labels add little predictive information for these six protein markers in spatially blocked validation.

\FloatBarrier
\section{Limitations}
\label{app:limitations}
\textsc{SwarmRSI} requires an evaluator that can execute candidate programs, so research without executable feedback is outside its scope.
Execution feedback does not always reveal whether the intended intervention was faithfully implemented, and performance gains alone do not establish its scientific mechanism.
Rebase selects program bases by matching recorded implementation needs to program mechanism labels, a heuristic that does not measure what a program can do.
Crossover applies only when a benchmark reports scores per instance.
H/P selection adapts discounted UCB without inheriting its guarantees, and its benefit is established only empirically.

Evaluation covers comparisons under matched budgets on three benchmark families.
The human failure analysis characterizes 644 annotated failed attempts from 35 task trajectories.
Cross-task evidence comes from six source tasks, 12 held-out tasks and one backbone, and generations are confounded with the number of source tasks seen.

The evaluation does not separately assess literature synthesis~\citep{shi2025deep}, idea quality~\citep{sauter2026evoideator}, or peer-review quality~\citep{guo2026rubricreviewer}.
Generalization to complete research workflows~\citep{lu2024ai,schmidgall2025agentlaboratory,evoscientist2026} and information-access agents~\citep{kanoulas2025agent,cao2026searchjev} requires additional evaluation.

Other untested settings include legal reasoning~\citep{lyu2022ceen,lyu2023multidefendant,wu2023fuzi}, zero-shot entity extraction~\citep{wang2025cmas}, conversational recommendation~\citep{zhang2024empathetic}, and shopping research~\citep{lyu2025deepshop}.
These settings require task-specific objectives and evaluation protocols.

\section{Ethics Statement}
\label{app:ethics}
Benchmark gains do not establish factual reliability or alignment with human preferences, which require separate evaluation and optimization~\citep{lin2022truthfulqa,lyu2024knowtuning,rafailov2023dpo,lyu2025macpo}.
We do not assess robustness to dataset shortcuts~\citep{lyu2023dct}, social stereotypes~\citep{parrish2022bbq}, or cognitive biases~\citep{lyu2024cognitivebiasesnews,lyu2025sacd}, nor do we test harmful-content generation.
Applications that affect people should preserve user agency and provide domain-appropriate human oversight~\citep{zhang20251st}.

Generated programs and scientific interpretations require independent validation before use beyond the benchmark setting.
The execution and scientific audits check task validity and implementation evidence; they do not certify security or scientific correctness.
The biological case studies analyse existing datasets, and their associations do not establish causal mechanisms or clinical validity.
Any deployment should restrict code execution, external-tool access and credentials to the approved scope, and review proposed uses for potential harm.

Data reuse should respect source licences, access restrictions and privacy requirements, with provenance and intended uses documented~\citep{gebru2021datasheets}.
We report time budgets, hardware configurations and token consumption to support resource-aware comparisons~\citep{schwartz2020greenai}.
Token consumption is not a direct measurement of energy use or carbon emissions.

\end{document}